\documentclass{article}
\usepackage{iclr2027_conference,times}

\usepackage[T1]{fontenc}
\usepackage{microtype}
\usepackage{xspace}
\usepackage{amsmath,amssymb,amsfonts}
\usepackage{mathtools}
\usepackage{booktabs}
\usepackage{multirow}
\usepackage{placeins}
\usepackage{float}
\usepackage{graphicx}
\usepackage{xcolor}
\usepackage{colortbl}
\usepackage{algorithm}
\usepackage{algorithmic}
\usepackage{url}
\usepackage{hyperref}
\usepackage[nameinlink,capitalise,noabbrev]{cleveref}

\hypersetup{
    colorlinks=true,
    linkcolor=blue,
    citecolor=blue,
    urlcolor=blue,
    pdftitle={RefAdapt-DiT: Adaptive Joint Attention for Reference-Conditioned Diffusion Transformers},
    pdfauthor={Jian Tang, Jiawei Fan, Qiannan Zhou, Qingbin Liu, Jiang Bian, Zang Li},
    pdfkeywords={diffusion transformer, reference-conditioned generation, joint attention, inference acceleration, KV cache}
}

\newcommand{\RefTok}{\mathrm{R}}
\newcommand{\TgtTok}{\mathrm{Y}}
\newcommand{\softmax}{\operatorname{Softmax}}

\newcommand{\norm}[1]{\left\lVert #1 \right\rVert}
\newcommand{\eps}{\varepsilon}
\newcommand{\RefAdapt}{\textsc{RefAdapt-DiT}\xspace}
\definecolor{tabgray}{gray}{0.955}
\newcommand{\qgain}[1]{{\scriptsize\textcolor{green!45!black}{(#1)}}}
\newcommand{\qdrop}[1]{{\scriptsize\textcolor{red!70!black}{(#1)}}}
\newcommand{\qflat}[1]{{\scriptsize\textcolor{gray!70}{(#1)}}}
\newcommand{\deltacell}[2]{\shortstack{#1\\[-1pt]#2}}

\title{RefAdapt-DiT: Adaptive Joint Attention\\for Reference-Conditioned\\Diffusion Transformers}

\iclrfinalcopy
\author{%
  Jian Tang\textsuperscript{1}\thanks{Corresponding author.} \quad
  Jiawei Fan\textsuperscript{1} \quad
  Qiannan Zhou\textsuperscript{1} \quad
  Qingbin Liu\textsuperscript{1} \quad
  Jiang Bian\textsuperscript{1} \quad
  Zang Li\textsuperscript{1} \\[0.5em]
  \textsuperscript{1}Platform and Content Group, Tencent \\
  \texttt{jackyjtang@tencent.com}
}
\date{}

\begin{document}
\raggedbottom
\maketitle
\lhead{Preprint}

\begin{abstract}
Diffusion Transformers (DiTs) have become the standard backbone for high-quality generative modeling, yet deploying them in conditional generation tasks remains computationally prohibitive because bidirectional joint attention repeatedly processes large reference streams. While existing optimization schemes mitigate generic temporal redundancy, they typically rely on coarse-grained static reuse and overlook the distinct dynamics of references and targets. Specifically, we observe that reference representations often evolve slowly along the generation trajectory, while the target often assigns little attention mass to them; reference drift and this target-to-reference exposure jointly shape how strongly stale reference states affect the target. To exploit these patterns, we introduce \RefAdapt, a training-free framework for adaptive control of joint attention between references and targets. Instead of rigid static strategies, \RefAdapt combines consecutive target-Q change with previously observed target-to-reference attention mass to control reference computation adaptively at block granularity. Under ultra-few-step settings, \RefAdapt enables speedups of up to $2.097\times$ on 4-step MiniMax H3 and $3.54\times$ on 8-step Qwen Image Edit, while maintaining comparable visual quality.
\end{abstract}


\section{Introduction}
\label{sec:intro}

Diffusion Transformers (DiTs) are now a common backbone for high-fidelity image
and video generation \citep{dit,qwenimage}. Reference-conditioned systems
increasingly use joint attention, placing visual references, instructions, and
noisy target tokens in a shared Transformer sequence
\citep{omnigen,unireal,unic,editverse,videoasprompt}. As these systems expand
from single-reference editing to multi-reference image generation and
temporally extended video editing \citep{multibanana,unic,editverse}, the
reference stream can occupy a large fraction of the sequence. Its cost is paid
repeatedly during denoising: although the external reference is fixed, its
hidden states and attention K/V continue to evolve through bidirectional
interaction with the target.

Training-free acceleration exploits redundancy either across denoising steps,
through caching or forecasting \citep{teacache,seacache,taylorseer,rfc,hyca},
or within each step, through token and attention reduction
\citep{tome,asymrnr,vmoba}. Reference-side methods further compress or reuse
conditioning features, alter the conditioning dependency, or exploit localized
editing redundancy \citep{ominicontrol2,easycontrol,regione}. These methods
show that substantial computation can be removed, but they leave a distinct
question when the original bidirectional joint attention is preserved: a cached
reference can be redundant or predictable yet still be important to the current
target. Reference reuse therefore requires estimating not only how stale the
reference is, but also the consequence of that staleness for the target.

We first use full-length, full-fresh trajectories to expose a real but
nonuniform reference-reuse opportunity; their dense adjacent transitions provide
temporal resolution for mechanism analysis. Across H3 video generation and Qwen
multi-reference image editing, reference K/V
evolves more slowly than the target through broad shallow-to-middle depth
regimes, while this asymmetry narrows and can reverse in deeper blocks. This
role-dependent temporal redundancy motivates reuse, but does not by itself make
reuse safe. The target reads reference tokens with strongly varying directional
exposure across depth and denoising steps, and controlled stale-K/V
counterfactuals show that the resulting target perturbation depends on this
interaction. Conditioning reference drift on reference-read exposure raises
its Spearman association with local target-output impact from $0.564$ to $0.826$
in H3 and from $0.708$ to $0.905$ in Qwen. The key distinction is therefore
between \emph{reference motion} and the \emph{target consequence} of that motion.

This distinction creates the online design problem. The counterfactual analysis
can measure fresh reference drift and stale-reference impact offline, but neither
quantity is available before a reuse decision without performing the computation
we aim to avoid. \RefAdapt therefore anticipates target consequence from two
signals already available from the target path and its recent history: current
target-Q change and previously observed directional exposure. The resulting
block-local score is evaluated from shallow to deep. Once it exceeds a threshold,
\RefAdapt reactivates the cached reference hidden state at that depth and
recomputes the remaining reference suffix; shallower blocks continue to use
cached reference K/V. Because the policy adapts reference depth rather than token
count or generic denoiser reuse, it can be composed with those acceleration
mechanisms; we evaluate such combinations directly rather than assuming
additive gains.

We evaluate \RefAdapt on 100 MultiBanana cases with Qwen Image Edit and 100
EditVerseBench cases with MiniMax H3, using benchmark-native quality metrics and
measured generation efficiency. Our contributions are:
\begin{itemize}
    \item We identify a depth-dependent reference-reuse opportunity in
    joint-attention diffusion: reference K/V changes substantially more slowly
    than the target in broad shallow-to-middle regimes, but not uniformly across
    depth.
    \item We show with controlled stale-reference counterfactuals that reference
    staleness is interaction-conditioned: directional target exposure makes
    reference drift substantially more predictive of target-side attention
    error.
    \item We introduce \RefAdapt, an impact-aware depth-adaptive reuse policy that
    combines current target-Q dynamics with lagged directional exposure, then
    reactivates reference computation at a calibrated depth cutoff without
    retraining.
\end{itemize}

\section{Related Work}
\label{sec:related}

\paragraph{Reference-conditioned joint-attention generation.}
Unified diffusion models formulate editing and reference-guided generation as
in-context generation, processing source/reference content and noisy target
tokens in a shared Transformer computation
\citep{omnigen,unireal,dreamomni,multibanana,unic,editverse,videoasprompt}.
This design supports multi-reference images and video context, but repeatedly
updates the reference stream during denoising. We study efficiency within this
existing bidirectional joint-attention formulation rather than introducing a
new conditioning architecture.

\paragraph{Training-free diffusion acceleration.}
Temporal reuse methods cache intermediate activations or adapt reuse using
trajectory evolution, forecasting, module relations, spectral structure, or
block/subspace dynamics
\citep{deepcache,cacheme,pab,fastercache,teacache,adacache,omnicache,dicache,taylorseer,rfc,hyca,svdcache,scalingcache,bwcache,seacache}.
A complementary family reduces within-step work through token merging or
attention reduction \citep{tome,asymrnr,astraea,vmoba,jenga}. These methods
primarily estimate generic computational redundancy---whether a state, block,
or token can be reused, predicted, or removed. \RefAdapt instead asks whether
stale \emph{reference} information is exposed to and consequential for the
current target. Since this criterion controls reference refresh rather than
whole-denoiser reuse or token count, the mechanisms can be combined; we report
composed variants in the main comparison.

\paragraph{Efficient reference-side computation.}
Closest to our setting, OminiControl2 compresses conditional tokens and reuses
conditional features, while EasyControl changes the conditioning dependency so
conditional K/V can be reused \citep{ominicontrol2,easycontrol}. RegionE
focuses on instruction-based single-image editing and exploits spatially
localized edited/unedited trajectories together with region-instruction caching
\citep{regione}. In contrast, \RefAdapt preserves the original bidirectional
joint-attention computation and adapts \emph{when} reference states are
refreshed according to their predicted target impact. This refresh criterion
applies to both multi-reference image editing and reference-conditioned video
generation without making the reference branch independent of the target.


\section{From Reference Dynamics to Target-Side Staleness Impact}
\label{sec:observations}

\paragraph{Setting.}
We ask three linked questions: where reference states change slowly, how target
queries read those reusable rows, and whether exposure-conditioned staleness
better predicts the resulting local target-attention perturbation. We group each
model's native token partition into two operational roles: $\RefTok$ contains
reusable cached reference-conditioning rows, and $\TgtTok$ the remaining active-
generation rows. With $\tau_i\in\{\tau_{\mathrm{ref}},\tau_{\mathrm{target}}\}$,
the row sets are
\[
\RefTok=\{i\mid\tau_i=\tau_{\mathrm{ref}}\},\qquad
\TgtTok=\{i\mid\tau_i=\tau_{\mathrm{target}}\}.
\]
The superscript $\TgtTok\!\rightarrow\!\RefTok$ means target queries look up
reference keys (exposure, not causal flow). Here $t$ indexes denoising steps
and $b\in\{0,\ldots,B-1\}$ Transformer blocks. Sections~\ref{sec:obs-dynamics}
and~\ref{sec:obs-exposure} use full-length, full-fresh trajectories for temporal
resolution; \cref{sec:obs-utility} uses an earlier anchor for a read-only
counterfactual. These are mechanism diagnostics; distilled few-step deployment
is evaluated in \cref{sec:experiments}. Except for the three-prompt profiles in
\cref{fig:obs-dynamics}, all are within-run diagnostics.

\subsection{Reference--Target Dynamics: Potential Reuse Opportunity}
\label{sec:obs-dynamics}

\begin{figure}[!ht]
    \centering
    \includegraphics[width=\linewidth]{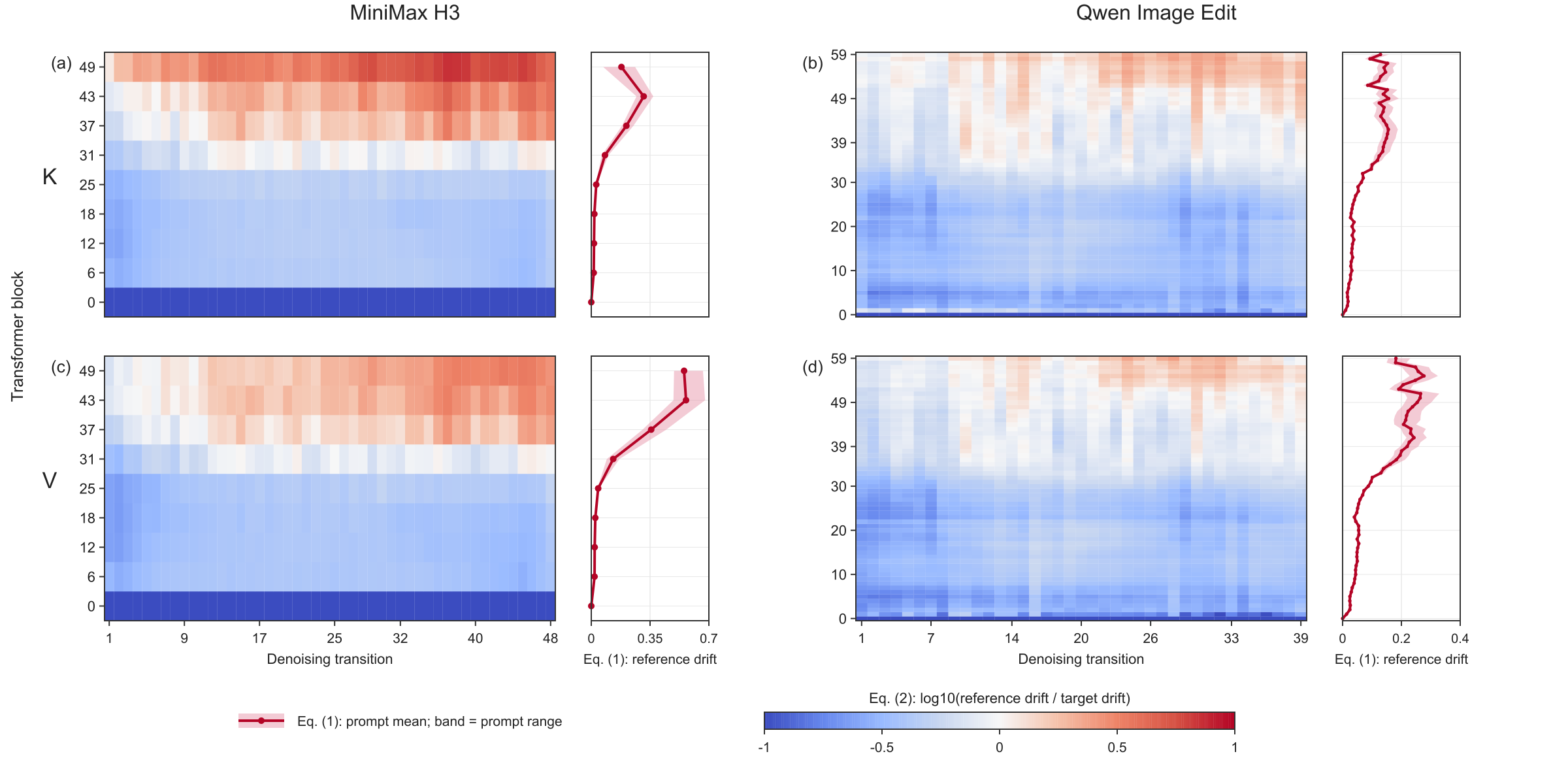}
    \caption{Matched reference--target K/V dynamics. Heatmaps show transition
    (horizontal), block (vertical), and prompt-averaged $\rho$ (color; negative
    means slower Ref motion). Right profiles show mean Ref drift horizontally
    and block vertically; pink spans prompts. H3 uses the nine shared blocks.}
    \label{fig:obs-dynamics}
\end{figure}

For any fixed-scope tensor slice $X$, define its consecutive relative
\emph{drift} as
\begin{equation}
D_{t,b}(X)=
\frac{\norm{X_{t,b}-X_{t-1,b}}_F}
{\norm{X_{t-1,b}}_F+\eps}.
\label{eq:consecutive-drift}
\end{equation}
where token scope and, when applicable, attention head are fixed; $X=Z^\tau$,
$Z\in\{K,V\}$, and $\tau\in\{\tau_{\mathrm{ref}},\tau_{\mathrm{target}}\}$.
Thus $D$ is adjacent-step motion, not cache error or anchor distance.
\Cref{fig:obs-dynamics} visualizes the matched log-ratio
\begin{equation}
\rho_{t,b}^{Z}=\log_{10}
\frac{D_{t,b}(Z^{\tau_{\mathrm{ref}}})+\eps}
{D_{t,b}(Z^{\tau_{\mathrm{target}}})+\eps},
\label{eq:drift-log-ratio}
\end{equation}
so $\rho<0$ means slower reference motion. In \cref{fig:obs-dynamics}, the
shallow-to-middle heatmaps are predominantly negative and the absolute Ref
profiles are small and flat. The median Ref/Tgt K/V ratio is $0.352$ in H3
blocks 6--25 and $0.433/0.381$ for Qwen K/V across depth. Because the comparison
is role matched, slower Ref motion indicates potential reuse, not cache safety;
the gap narrows or reverses deeper, making one depth-independent rule too coarse.

An H3 timing probe further finds median per-block Spearman $0.675$ between target
block-input change and Ref K/V change, with 22/50 blocks reaching $0.8$. Because
the tensors differ, this motivates a target-side runtime cue but neither
reconstructs Ref drift nor certifies reuse.

Each heatmap cell is the mean of $\rho_{t,b}^{Z}$ over $N$ prompts. The right
profiles retain absolute scale by averaging reference drift over transitions
within each prompt and then over prompts:
\begin{equation}
\bar D_{b}^{Z,\mathrm{ref}}=
\frac{1}{N}\sum_{n=1}^{N}
\left[\operatorname*{mean}_{t}
D_{t,b}^{(n)}(Z^{\tau_{\mathrm{ref}}})\right].
\label{eq:cross-prompt-reference-profile}
\end{equation}
where $n$ indexes prompts. The red curve is the mean and the pink envelope is
the prompt range; prompts are listed in \cref{app:fig1-prompts}.

\subsection{Directional Exposure: Target Queries Attending to Reference Keys}
\label{sec:obs-exposure}

Drift alone does not show how strongly the target reads reference rows. For
head $h$, let $\mathcal Q_{\TgtTok}$ be sampled target-query rows and
$\mathcal K_{\RefTok}$ reusable reference-key rows. For post-softmax attention
$A=\softmax(QK^\top/\sqrt{d_h})$, define their directional mass
\begin{equation}
M_{t,b,h}^{\TgtTok\rightarrow\RefTok}(A)=
\frac{1}{|\mathcal Q_{\TgtTok}|}
\sum_{q\in\mathcal Q_{\TgtTok}}
\sum_{k\in\mathcal K_{\RefTok}}
A_{qk}^{t,b,h}.
\label{eq:directional-exposure}
\end{equation}
Thus $M$ measures how much target attention is routed through reference values:
it is exposure, not by itself a causal effect or reuse criterion.
$A^{\mathrm{fresh}}$ uses current Q/K over all keys; the runtime controller
applies the same definition to executed attention.

\begin{figure}[!t]
    \centering
    \includegraphics[
        width=0.90\linewidth,
        trim=30bp 45bp 110bp 20bp,
        clip
    ]{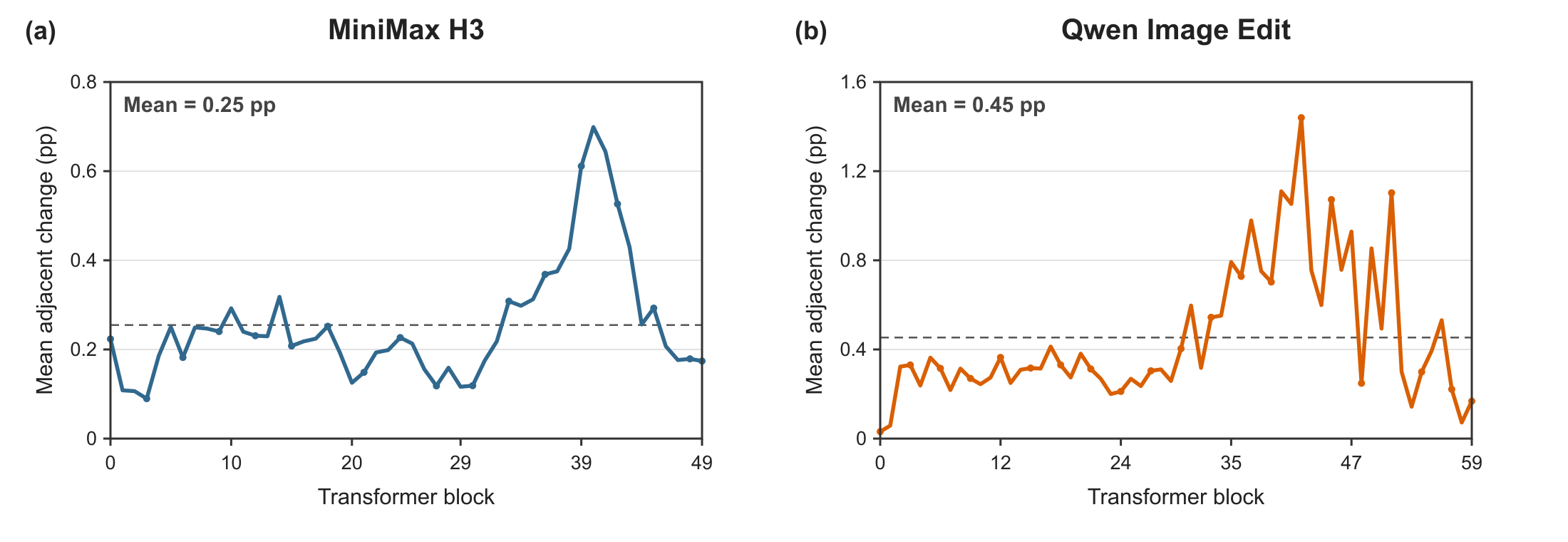}
    \caption{Adjacent-step directional Ref-mass stability. Horizontal: block;
    vertical: mean change $\bar\Delta_b$ in percentage points (shared scale).
    Mass is exposure, not causal effect or a standalone reuse rule.}
    \label{fig:obs-exposure}
\end{figure}

For the summaries in \cref{tab:obs-exposure-allocation,fig:obs-exposure}, we
average the per-head quantity,
\begin{equation}
\bar M_{t,b}^{\TgtTok\rightarrow\RefTok}
=\frac{1}{H}\sum_{h=1}^{H}
M_{t,b,h}^{\TgtTok\rightarrow\RefTok}(A^{\mathrm{fresh}}).
\label{eq:head-mean-exposure}
\end{equation}
Trajectory means in \cref{tab:obs-exposure-allocation} are $9.51\%$ Ref mass
for H3 and $14.66\%$ for Qwen; these describe exposure, not reuse thresholds.

To test adjacent-step similarity directly, we average the directional Ref-mass
change between neighboring steps at each fixed block:
\begin{equation}
\bar\Delta_{b}^{\TgtTok\rightarrow\RefTok}
\coloneqq \frac{100}{T-1}\sum_{t=2}^{T}\left\lvert
\bar M_{t,b}^{\TgtTok\rightarrow\RefTok}
-\bar M_{t-1,b}^{\TgtTok\rightarrow\RefTok}
\right\rvert
\quad\text{(percentage points)}.
\label{eq:adjacent-ref-change}
\end{equation}
\Cref{fig:obs-exposure} shows that $\bar\Delta_b$ is at most $0.70$ points in
H3 and $1.44$ in Qwen: blockwise Ref mass is stable on average, not guaranteed
for every head or transition.

\begin{table}[!t]
    \centering
    \caption{Trajectory-wide directional attention allocation in full-fresh joint attention.}
    \label{tab:obs-exposure-allocation}
    \small
    \setlength{\tabcolsep}{8pt}
    \renewcommand{\arraystretch}{1.04}
    \begin{tabular}{@{}lccc@{}}
        \toprule
        Model (steps $\times$ blocks) & Ref mass & Tgt mass & Tgt/Ref \\
        \midrule
        MiniMax H3 ($49\times50$) & 9.51\% & 90.49\% & $9.52\times$ \\
        Qwen Image Edit ($40\times60$) & 14.66\% & 85.34\% & $5.82\times$ \\
        \bottomrule
    \end{tabular}
\end{table}

\vspace{-0.35em}

\subsection{Counterfactual Local Perturbation from Stale Reference K/V}
\label{sec:obs-utility}

Let $a<t$ be an earlier synthetic anchor in a fixed three-step schedule. For
reference tensor $Z\in\{K,V\}$, define its anchor-to-current endpoint drift
\begin{equation}
D_{t\mid a,b}(Z^{\RefTok})=
\frac{\norm{Z^{\RefTok}_{t,b}-Z^{\RefTok}_{a,b}}_F}
{\norm{Z^{\RefTok}_{a,b}}_F+\eps}.
\label{eq:endpoint-drift}
\end{equation}
Unlike \cref{eq:consecutive-drift}, this is anchor-to-current, not adjacent-step,
change. Applying this definition separately to head-resolved K and V slices
yields their joint endpoint scalar $E^{KV}_{t\mid a,b,h}$; this diagnostic-only
quantity is not used by the deployed controller.

\begin{figure}[H]
    \centering
    \begin{minipage}[c]{0.42\linewidth}
        \centering
        \includegraphics[width=\linewidth]{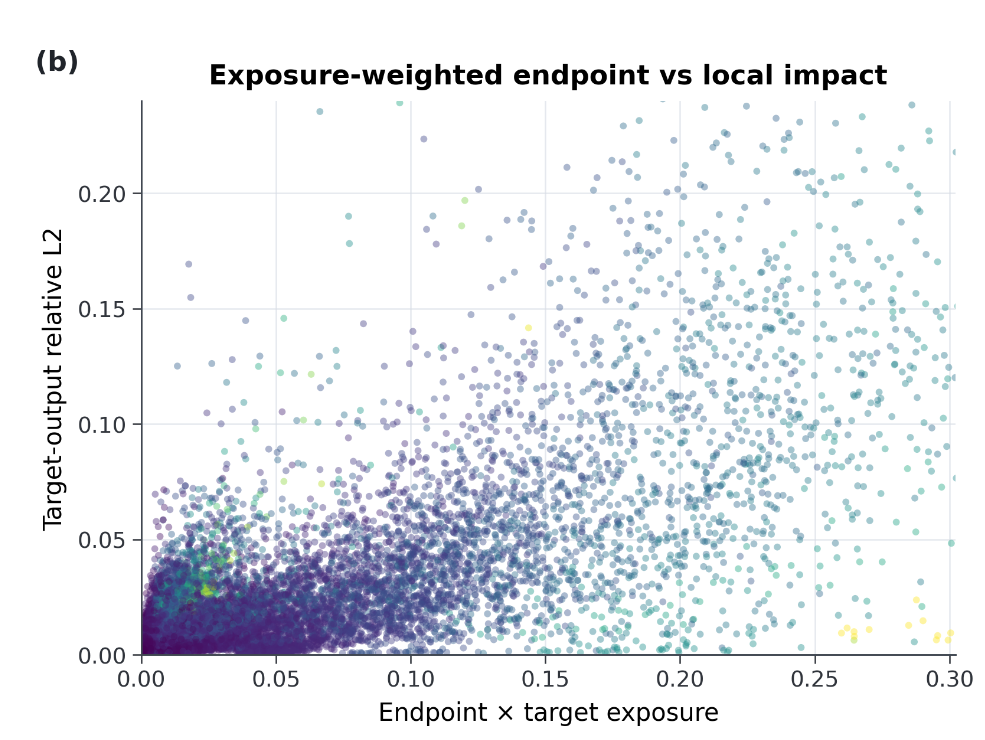}\\[-1pt]
        {\small (a) MiniMax H3}
    \end{minipage}\hfill
    \begin{minipage}[c]{0.42\linewidth}
        \centering
        \includegraphics[width=\linewidth]{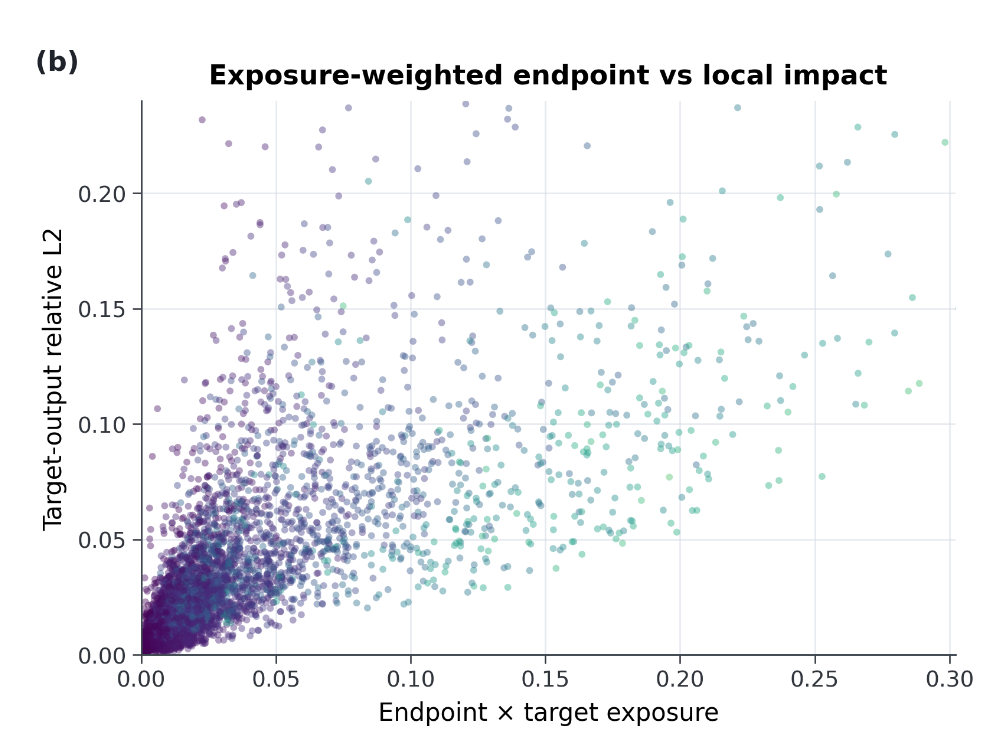}\\[-1pt]
        {\small (b) Qwen Image Edit}
    \end{minipage}\hfill
    \begin{minipage}[c]{0.07\linewidth}
        \centering
        \includegraphics[width=0.9\linewidth,height=1.58in,keepaspectratio]{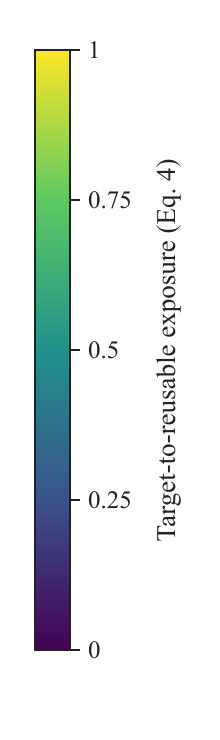}
    \end{minipage}
    \caption{Exposure-conditioned endpoint score versus local perturbation.
    Each point is one $(t,b,h)$ comparison: horizontal
    $E^{KV}M^{\TgtTok\rightarrow\RefTok}$, vertical $I^{KV}$, and color current
    per-head exposure $M$ to the operational Ref set defined above.
    This diagnostic does not define the deployed controller score.}
    \label{fig:obs-utility}
\end{figure}

Next, let $\operatorname{Attn}(Q,K,V)=\softmax(QK^\top/\sqrt{d_h})V$. Holding
current target queries and target-side K/V fixed, replace only reference K/V by
their anchor values:
\begin{align}
O_{t,b,h}^{\mathrm{fresh}}
&=\operatorname{Attn}\!\left(
Q_{\TgtTok}^{t,b,h},
[K_{\RefTok}^{t,b,h};K_{\TgtTok}^{t,b,h}],
[V_{\RefTok}^{t,b,h};V_{\TgtTok}^{t,b,h}]
\right),\\
O_{t\mid a,b,h}^{\mathrm{staleKV}}
&=\operatorname{Attn}\!\left(
Q_{\TgtTok}^{t,b,h},
[K_{\RefTok}^{a,b,h};K_{\TgtTok}^{t,b,h}],
[V_{\RefTok}^{a,b,h};V_{\TgtTok}^{t,b,h}]
\right).
\label{eq:fresh-stale-output}
\end{align}
The resulting local counterfactual perturbation is
\begin{equation}
I_{t\mid a,b,h}^{KV}=
\frac{\norm{O_{t\mid a,b,h}^{\mathrm{staleKV}}-
O_{t,b,h}^{\mathrm{fresh}}}_F}
{\norm{O_{t,b,h}^{\mathrm{fresh}}}_F+\eps}.
\label{eq:local-impact}
\end{equation}
In \cref{fig:obs-utility}, each $(t,b,h)$ point has horizontal coordinate
$E^{KV}M(A^{\mathrm{fresh}})$, vertical coordinate $I^{KV}$, and color equal
to the same exposure $M$. The Frobenius ratio is relative L2 on the target
attention output; the stale branch is read-only, so $I^{KV}$ is neither a
perceptual metric nor a final-error bound.

Raw $E^{KV}$ has Spearman $0.564/0.708$ with $I^{KV}$ on H3/Qwen; exposure
weighting raises this to $0.826/0.905$ in every measured block. Thus equal Ref
drift can have different local effects depending on target exposure. The
product is a diagnostic, not an error model or calibrated probability.
\FloatBarrier


\section{RefAdapt-DiT: Depth-Adaptive Reference Reuse}
\label{sec:method}

The probes above suggest that reuse should be depth aware and conditioned on
target--reference interaction. Because fresh Ref drift is unavailable during
reuse, \RefAdapt combines observable target-Q change with recent directional
exposure to choose a depth cutoff; a bounded-age refresh limits staleness.
\Cref{fig:method-overview} separates the cached state, online cues, and action,
because each block decision must be made before current Ref rows exist.

\begingroup
\setlength{\intextsep}{4pt plus 1pt minus 1pt}
\begin{figure}[H]
    \centering
    \includegraphics[
        width=\linewidth,
        trim=0bp 40bp 0bp 0bp,
        clip
    ]{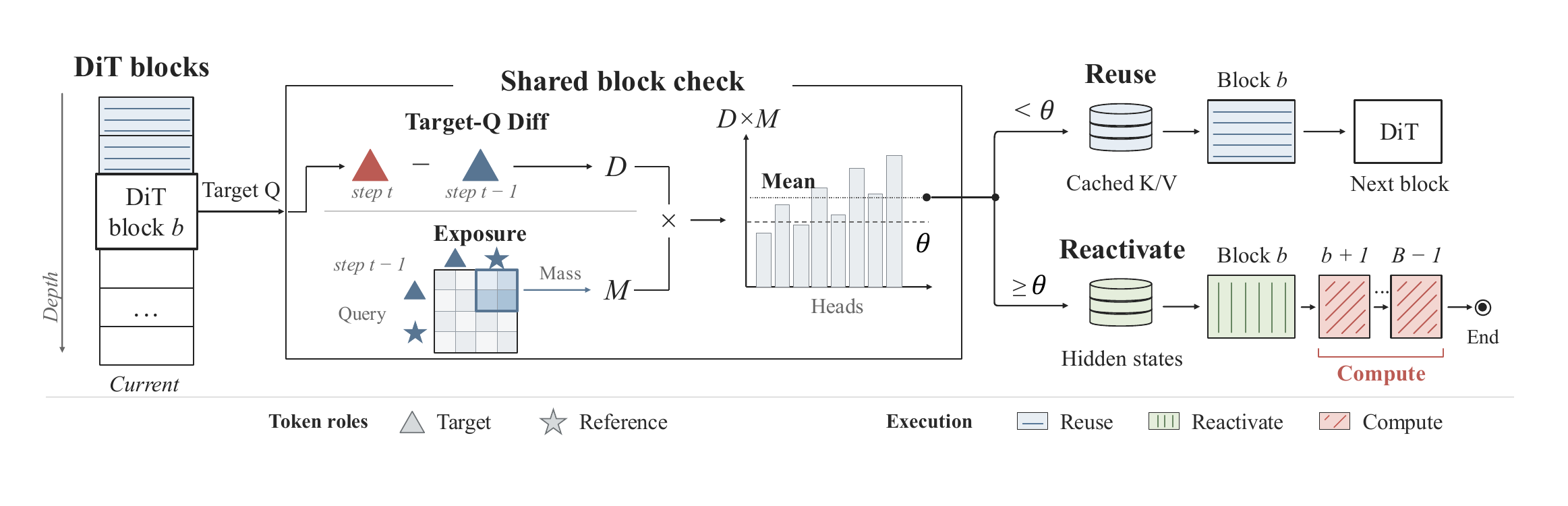}
    \caption{\RefAdapt at one step. Per-head target-Q change $D$ and lagged
    reference exposure $M$ form $D\times M$; the first block whose head mean
    reaches $\theta$ sets the depth cutoff. Earlier blocks \textsc{reuse},
    the boundary \textsc{reactivate}s, and deeper blocks \textsc{compute}.}
    \label{fig:method-overview}
\end{figure}
\endgroup

\paragraph{Execution and cache state.}
A \emph{reuse} block skips reference rows while the target attends to cached
reference K/V; \emph{reactivation} restores a cached boundary state before
deeper blocks compute and update the reference stream. A \emph{full refresh}
starts at block 0. Each block $b$ stores
\begin{equation}
\mathcal Z_b=
\left(
\bar H_b^{\RefTok},
\bar K_b^{\RefTok},
\bar V_b^{\RefTok},
a_b
\right),
\label{eq:block-cache-state}
\end{equation}
where $\bar H_b^{\RefTok}$ is the cached block-input state,
$(\bar K_b^{\RefTok},\bar V_b^{\RefTok})$ its K/V, and $a_b$ its last
materialization step. During reuse, cached K/V serve target attention without
propagating a new Ref output. Restoring $\bar H_b^{\RefTok}$ at a boundary gives
deeper blocks a coherent Ref stream; computing only a suffix therefore yields
depth-dependent cache ages without disconnected per-block refreshes.

\paragraph{Runtime cues observable under reuse.}
\label{sec:method-cues}

Fresh-reference diagnostics are unavailable under reuse, but target Q remains
active. The timing probe in \cref{sec:obs-dynamics} motivates this target-side
cue without treating it as reconstructed Ref drift. We apply
\cref{eq:consecutive-drift} to each post-RoPE target-query head, obtaining
$D_{t,b}(Q^{\TgtTok,h})$. Directional interaction uses
\cref{eq:directional-exposure} on executed attention $A^{\mathrm{run}}$.
Because block $b$ has not run when deciding step $t$, the available cue is
$M^{\TgtTok\rightarrow\RefTok}_{t-1,b,h}(A^{\mathrm{run}})$. It is recorded
without another attention evaluation; a full step 0 initializes all state.
Thus target-Q change supplies current dynamics. The adjacent-step stability in
\cref{sec:obs-exposure} supports treating the last executed mass as a recent
interaction cue when current Ref rows are unavailable, not as an estimate of
current fresh attention.

\paragraph{Interaction-conditioned score.}
\label{sec:method-score}

Target-Q drift measures current change, while lagged mass measures recent Ref
interaction. Their zero-parameter product transfers the diagnostic's
change--interaction factorization into the per-head reactivation score
\begin{equation}
g_{t,b,h}
=
D_{t,b}(Q^{\TgtTok,h})
\cdot
M^{\TgtTok\rightarrow\RefTok}_{t-1,b,h}(A^{\mathrm{run}}).
\label{eq:factorized-risk}
\end{equation}
It is small if either factor is weak and controls reactivation; it neither
reconstructs unavailable fresh Ref drift/current exposure nor estimates
counterfactual error.
The product is formed per head before aggregation, preserving whether change
and exposure coincide within the same head.

\paragraph{Depth cutoff and bounded-age refresh.}
\label{sec:method-action}

\RefAdapt scans shallow to deep. The first block whose head-mean score reaches
$\theta$ is the reactivation boundary $b_t$:
\begin{equation}
\frac{1}{H}\sum_{h=1}^{H} g_{t,b_t,h} \ge \theta,
\qquad
b_t \text{ is the first block satisfying this condition.}
\label{eq:depth-cutoff}
\end{equation}
If none crosses, the entire step reuses. Before $b_t$, blocks use cached K/V; at
$b_t$, the target still uses cached K/V while $\bar H_{b_t}^{\RefTok}$ is
restored, so new Ref K/V begin at $b_t+1$ and the deeper suffix recomputes. After
$\Delta_{\max}$ steps without a full refresh, the controller recomputes from
block 0, updates all caches, and resets their ages. The first-crossing rule keeps
reference computation as one contiguous suffix with a single restored boundary,
avoiding incompatible independent block decisions.

\section{Experiments}
\label{sec:experiments}

\vspace{-0.35em}
\subsection{Experimental Setup}
\vspace{-0.20em}
\label{sec:exp-setup}

We evaluate the same training-free controller design on fixed 100-case Qwen
Image Edit~\citep{qwenimage}/MultiBanana~\citep{multibanana} image and MiniMax
H3/EditVerseBench~\citep{editverse} video sets, with model, sampler, precision,
resolution, batch size, and hardware fixed within each setting. Baselines are
TeaCache, SeaCache, TaylorSeer, ToMe, and H3-only AsymRnR.

We use native metrics. MultiBanana reports IA, RC, BSM, PR, VQ, and official
score $3\,\mathrm{IA}+3\,\mathrm{RC}+\mathrm{BSM}+\mathrm{PR}+\mathrm{VQ}$;
EditVerseBench reports EditQ, Pick Score, frame/video text alignment, and
CLIP/DINO temporal consistency. The main experiments use distilled few-step
checkpoints on NVIDIA H20 GPUs with 96 GB memory per GPU: distilled 8-step Qwen
Image Edit at $1024\times1024$ on one GPU, and the LightX2V release of distilled
4-step MiniMax H3 at $1344\times768$ (768p) on the same GPU type. Reported
speedup uses measured generation latency relative to full computation; all
timings come from real executions rather than offline schedule simulation.
Unlike the full-length diagnostics, these
few-step settings provide a stricter error-recovery test, with fewer later steps
to correct a reuse error; full protocol details are in \cref{app:experiments}.
All methods use the same cases and setting-specific configuration. Composed
ResComp rows are reported separately so standalone reuse gains remain explicit.

\vspace{-0.35em}
\subsection{Main Comparison}
\vspace{-0.20em}
\label{sec:exp-main}

\paragraph{Qwen image editing.}

\Cref{tab:image-main} reports all MultiBanana metrics; TaylorSeer uses interval
2, and the composed row adds orthogonal resolution compression to \RefAdapt.
Colored parentheses in both main tables show changes from full computation.
The Type column uses C/F/T/R/H for caching, forecasting, token reduction,
reference reuse, and hybrid methods, respectively.

\begingroup
\setlength{\intextsep}{6pt plus 1pt minus 1pt}
\begin{table}[H]
\caption{Qwen Image Edit on 100 MultiBanana cases (8 steps). MB Score is the
official aggregate; time is in seconds. \textbf{ResComp} denotes orthogonal
resolution compression.}
\label{tab:image-main}
\centering
\footnotesize
\setlength{\tabcolsep}{1.65pt}
\renewcommand{\arraystretch}{1.00}
\resizebox{\textwidth}{!}{%
\begin{tabular}{@{}llrlrrrrrr@{}}
\toprule
\multirow{2}{*}{\textbf{Method}} &
\multirow{2}{*}{\textbf{Type}} &
\multicolumn{2}{c}{\textbf{Efficiency}} &
\multicolumn{6}{c}{\textbf{MultiBanana Evaluation}} \\
\cmidrule(lr){3-4} \cmidrule(l){5-10}
& & \textbf{Speedup $\uparrow$} & \textbf{Time $\downarrow$}
& IA $\uparrow$ & RC $\uparrow$ & BSM $\uparrow$
& PR $\uparrow$ & VQ $\uparrow$
& \textbf{MB Score $\uparrow$} \\
\midrule
\rowcolor{tabgray}
Full computation
& -- & 1.00$\times$ & 60.39
& 1.286 & 1.735 & 1.408 & 1.878 & 2.163 & 14.512 \\

TeaCache~\citep{teacache}
& C & 1.13$\times$ & 53.30
& \deltacell{1.367}{\qgain{$+0.081$}}
& \deltacell{1.571}{\qdrop{$-0.164$}}
& \deltacell{1.306}{\qdrop{$-0.102$}}
& \deltacell{1.816}{\qdrop{$-0.062$}}
& \deltacell{2.041}{\qdrop{$-0.122$}}
& \deltacell{13.977}{\qdrop{$-0.535$}} \\

SeaCache~\citep{seacache}
& C & 1.60$\times$ & 37.63
& \deltacell{1.347}{\qgain{$+0.061$}}
& \deltacell{\textbf{1.959}}{\qgain{$+0.224$}}
& \deltacell{1.551}{\qgain{$+0.143$}}
& \deltacell{1.776}{\qdrop{$-0.102$}}
& \deltacell{2.122}{\qdrop{$-0.041$}}
& \deltacell{15.367}{\qgain{$+0.855$}} \\

TaylorSeer~\citep{taylorseer} ($I{=}2$)
& F & 1.61$\times$ & 37.58
& \deltacell{\textbf{1.396}}{\qgain{$+0.110$}}
& \deltacell{1.875}{\qgain{$+0.140$}}
& \deltacell{1.438}{\qgain{$+0.030$}}
& \deltacell{\textbf{2.083}}{\qgain{$+0.205$}}
& \deltacell{\textbf{2.271}}{\qgain{$+0.108$}}
& \deltacell{\textbf{15.605}}{\qgain{$+1.093$}} \\

ToMe~\citep{tome}
& T & 1.06$\times$ & 56.74
& \deltacell{1.200}{\qdrop{$-0.086$}}
& \deltacell{1.180}{\qdrop{$-0.555$}}
& \deltacell{1.120}{\qdrop{$-0.288$}}
& \deltacell{1.280}{\qdrop{$-0.598$}}
& \deltacell{1.180}{\qdrop{$-0.983$}}
& \deltacell{10.720}{\qdrop{$-3.792$}} \\

\addlinespace[1pt]
\midrule

\textbf{\RefAdapt}
& \textbf{R} & \textbf{2.58$\times$} & \textbf{23.43}
& \deltacell{1.320}{\qgain{$+0.034$}}
& \deltacell{1.640}{\qdrop{$-0.095$}}
& \deltacell{\textbf{1.620}}{\qgain{$+0.212$}}
& \deltacell{1.720}{\qdrop{$-0.158$}}
& \deltacell{2.220}{\qgain{$+0.057$}}
& \deltacell{14.440}{\qdrop{$-0.072$}} \\

\textbf{\RefAdapt + ResComp}
& \textbf{H} & \textbf{3.54$\times$} & \textbf{17.04}
& \deltacell{1.180}{\qdrop{$-0.106$}}
& \deltacell{1.180}{\qdrop{$-0.555$}}
& \deltacell{1.160}{\qdrop{$-0.248$}}
& \deltacell{1.920}{\qgain{$+0.042$}}
& \deltacell{2.160}{\qdrop{$-0.003$}}
& \deltacell{12.320}{\qdrop{$-2.192$}} \\

\bottomrule
\end{tabular}%
}
\end{table}
\endgroup

At 8 steps, \RefAdapt is the fastest standalone method in
\cref{tab:image-main}: $2.58\times$ speedup versus at most $1.61\times$ for the
baselines, with an MB Score close to full computation (14.440 versus 14.512).
Adding ResComp reaches $3.54\times$ but lowers the MB Score to 12.320, exposing
a separate, more aggressive operating point rather than conflating resolution
compression with the standalone controller gain.
The component scores make that tradeoff explicit: standalone \RefAdapt improves
IA and VQ over full computation and has the best BSM (1.620; next best 1.551),
while RC and PR decrease modestly. Thus
the near-matched aggregate does not hide a uniform degradation across all five
dimensions.
TaylorSeer attains a higher aggregate MB Score, but at $1.61\times$ rather than
$2.58\times$ speedup; \RefAdapt therefore occupies a distinct higher-speed
operating regime rather than dominating every quality dimension.

\begingroup
\setlength{\intextsep}{4pt}
\begin{figure}[H]
    \centering
    \includegraphics[
        width=0.84\textwidth,
        trim=110bp 190bp 122bp 178bp,
        clip
    ]{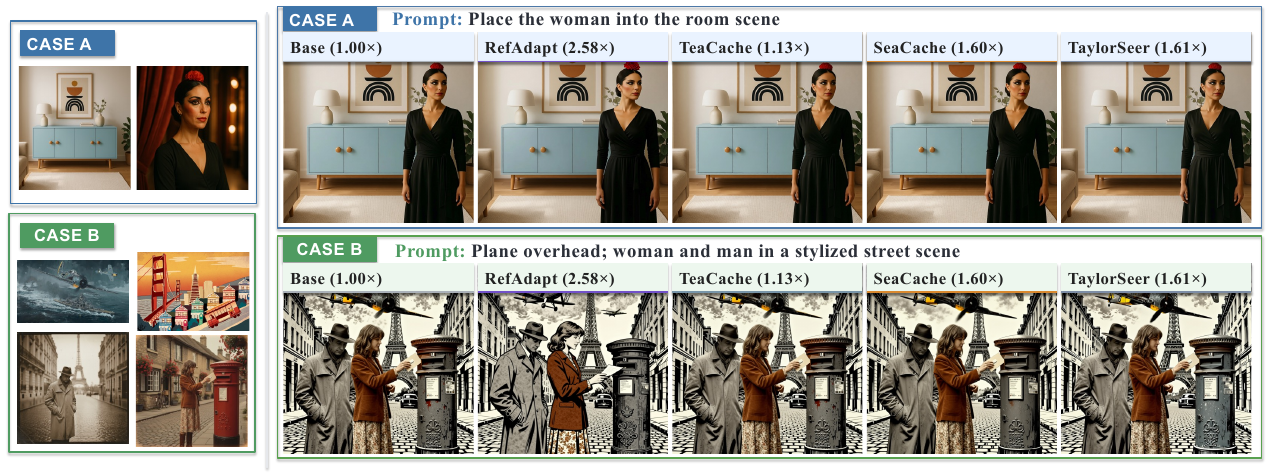}
    \caption{Qwen qualitative comparison for two multi-reference cases.
    Columns show full computation and four accelerated methods.}
    \label{fig:qwen-qualitative}
\end{figure}
\vspace{-0.35em}
\endgroup

In Case A of \cref{fig:qwen-qualitative}, the $2.58\times$ \RefAdapt output
preserves subject identity and room composition. In Case B, it appears to retain
the reference's line-art style more clearly than the lower-speed outputs shown,
while keeping both subjects and the aircraft.

\paragraph{MiniMax H3 video generation.}\mbox{}

\begingroup
\setlength{\intextsep}{4pt}
\begin{figure}[H]
    \centering
    \includegraphics[
        width=0.84\textwidth,
        trim=67bp 151bp 68bp 157bp,
        clip
    ]{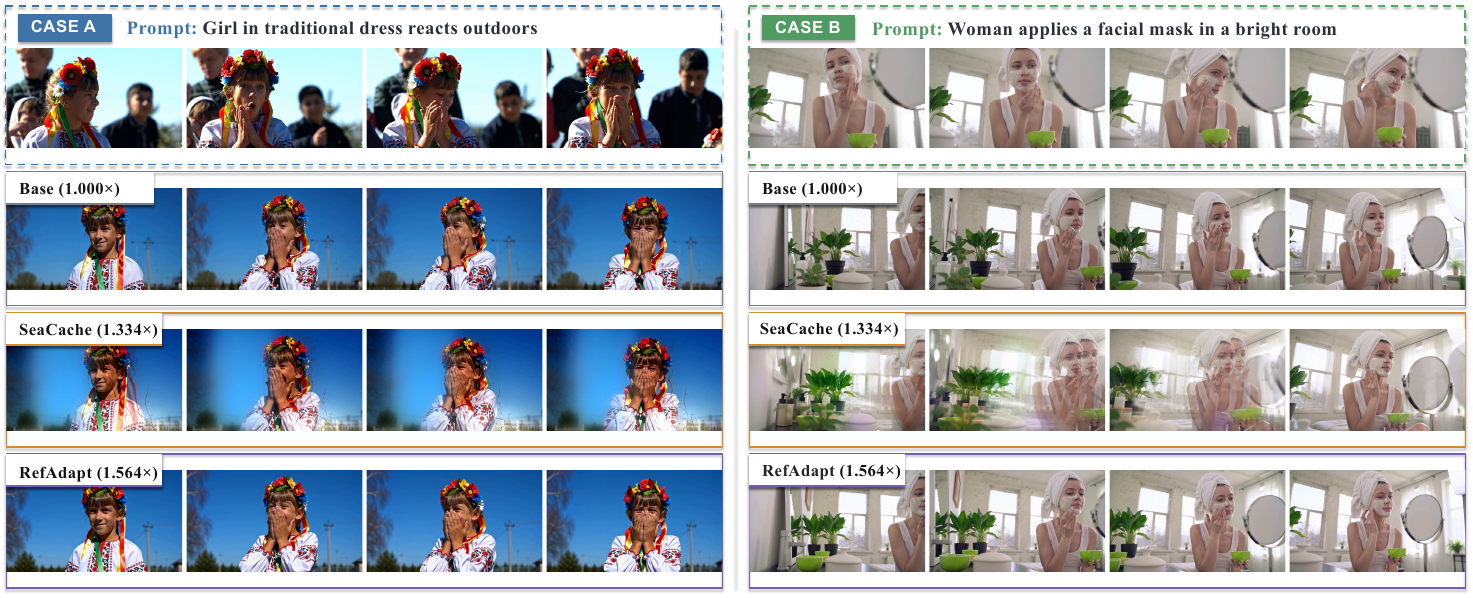}
    \caption{H3 qualitative comparison over sampled frames. Rows show full
    computation, SeaCache, and \RefAdapt.}
    \label{fig:h3-qualitative}
\end{figure}
\vspace{-0.35em}
\endgroup

\Cref{fig:h3-qualitative} shows \RefAdapt following Case A's motion without
freezing the trajectory while retaining reference appearance. In this sampled
Case B example, \RefAdapt keeps cleaner contours and scene structure while the
SeaCache output shows visible ghosting.

\begingroup
\setlength{\intextsep}{2pt}
\begin{table}[H]
\caption{MiniMax H3 on 100 EditVerseBench cases (4 steps). Quality follows the
official protocol; time is in seconds. \textbf{ResComp} denotes orthogonal
resolution compression.}
\label{tab:video-main}
\centering
\scriptsize
\setlength{\tabcolsep}{1.85pt}
\renewcommand{\arraystretch}{0.86}
\resizebox{\textwidth}{!}{%
\begin{tabular}{@{}llrlrrrrrr@{}}
\toprule
\multirow{2}{*}{\textbf{Method}} &
\multirow{2}{*}{\textbf{Type}} &
\multicolumn{2}{c}{\textbf{Efficiency}} &
\multicolumn{1}{c}{\textbf{VLM Eval.}} &
\multicolumn{1}{c}{\textbf{Video Qual.}} &
\multicolumn{2}{c}{\textbf{Text Align.}} &
\multicolumn{2}{c}{\textbf{Temporal Cons.}} \\
\cmidrule(lr){3-4}
\cmidrule(lr){5-5}
\cmidrule(lr){6-6}
\cmidrule(lr){7-8}
\cmidrule(l){9-10}
& & \textbf{Speedup $\uparrow$} & \textbf{Time $\downarrow$}
& \textbf{EditQ $\uparrow$}
& Pick $\uparrow$
& Frame $\uparrow$
& Video $\uparrow$
& CLIP $\uparrow$
& DINO $\uparrow$ \\
\midrule

\rowcolor{tabgray}
Full computation
& -- & 1.000$\times$ & 200.80
& 7.280 & \textbf{20.528} & 28.243 & 24.994 & 0.990 & \textbf{0.990} \\

TeaCache~\citep{teacache}
& C & 1.341$\times$ & 149.69
& \deltacell{7.340}{\qgain{$+0.060$}}
& \deltacell{20.509}{\qdrop{$-0.019$}}
& \deltacell{28.294}{\qgain{$+0.051$}}
& \deltacell{25.051}{\qgain{$+0.057$}}
& \deltacell{0.990}{\qflat{$+0.000$}}
& \deltacell{\textbf{0.990}}{\qflat{$+0.000$}} \\

SeaCache~\citep{seacache}
& C & 1.334$\times$ & 150.52
& \deltacell{7.060}{\qdrop{$-0.220$}}
& \deltacell{20.389}{\qdrop{$-0.139$}}
& \deltacell{28.197}{\qdrop{$-0.046$}}
& \deltacell{24.800}{\qdrop{$-0.194$}}
& \deltacell{0.988}{\qdrop{$-0.002$}}
& \deltacell{0.986}{\qdrop{$-0.004$}} \\

TaylorSeer~\citep{taylorseer}
& F & 1.333$\times$ & 150.67
& \deltacell{7.361}{\qgain{$+0.081$}}
& \deltacell{20.506}{\qdrop{$-0.022$}}
& \deltacell{28.305}{\qgain{$+0.062$}}
& \deltacell{25.062}{\qgain{$+0.068$}}
& \deltacell{0.990}{\qflat{$+0.000$}}
& \deltacell{\textbf{0.990}}{\qflat{$+0.000$}} \\

ToMe~\citep{tome}
& T & 1.207$\times$ & 166.37
& \deltacell{5.028}{\qdrop{$-2.252$}}
& \deltacell{19.851}{\qdrop{$-0.677$}}
& \deltacell{27.755}{\qdrop{$-0.488$}}
& \deltacell{24.614}{\qdrop{$-0.380$}}
& \deltacell{0.988}{\qdrop{$-0.002$}}
& \deltacell{0.985}{\qdrop{$-0.005$}} \\

AsymRnR~\citep{asymrnr}
& T & 1.044$\times$ & 192.29
& \deltacell{6.407}{\qdrop{$-0.873$}}
& \deltacell{20.196}{\qdrop{$-0.332$}}
& \deltacell{28.341}{\qgain{$+0.098$}}
& \deltacell{25.003}{\qgain{$+0.009$}}
& \deltacell{0.980}{\qdrop{$-0.010$}}
& \deltacell{0.977}{\qdrop{$-0.013$}} \\

\addlinespace[1pt]
\midrule

\textbf{\RefAdapt}
& \textbf{R} & \textbf{1.564$\times$} & \textbf{128.38}
& \deltacell{7.413}{\qgain{$+0.133$}}
& \deltacell{20.437}{\qdrop{$-0.091$}}
& \deltacell{28.053}{\qdrop{$-0.190$}}
& \deltacell{25.041}{\qgain{$+0.047$}}
& \deltacell{\textbf{0.991}}{\qflat{$+0.001$}}
& \deltacell{\textbf{0.990}}{\qflat{$+0.000$}} \\

\textbf{\RefAdapt + ResComp}
& \textbf{H} & \textbf{2.097$\times$} & \textbf{95.75}
& \deltacell{\textbf{7.565}}{\qgain{$+0.285$}}
& \deltacell{20.444}{\qdrop{$-0.084$}}
& \deltacell{\textbf{28.414}}{\qgain{$+0.171$}}
& \deltacell{\textbf{25.206}}{\qgain{$+0.212$}}
& \deltacell{\textbf{0.991}}{\qflat{$+0.001$}}
& \deltacell{\textbf{0.990}}{\qflat{$+0.000$}} \\

\bottomrule
\end{tabular}%
}
\vspace{-0.45em}
\end{table}
\endgroup

In \cref{tab:video-main}, on 4-step H3, \RefAdapt is the fastest standalone
method ($1.564\times$) and has the highest EditQ among standalone accelerators
(7.413 versus at most 7.361), while alignment and temporal consistency remain
near full computation. With ResComp it reaches $2.097\times$ while
retaining the overall quality profile, demonstrating that the reuse controller
composes with an orthogonal optimization even when only four denoising steps
are available.
Relative to TeaCache, the fastest cache baseline at $1.341\times$, standalone
\RefAdapt provides a 16.6\% larger speedup. CLIP/DINO temporal consistency
remains at
0.991/0.990; the small alignment shifts therefore do not indicate a broad
quality collapse.

\vspace{-0.35em}
\subsection{Ablation Studies}
\vspace{-0.20em}
\label{sec:exp-ablation}

\paragraph{Runtime-controller ablation.}

The 50-step Ref2VA trajectory is used only to expose enough decisions for signal
selection, not to estimate few-step deployment gain, and is separate from the
4-step main benchmark. We compare
consecutive target-Q change, directional exposure, their \RefAdapt product, and
anchor-relative target-Q change times exposure; Fixed R3 is a non-adaptive
baseline. All adaptive variants share the depth-wise executor in
\cref{sec:method}, the sequential stopping rule, bounded-age fallback,
threshold, and other settings. Only the trigger cue changes. Because this
executor can reuse a shallow prefix and recompute a deeper suffix within one
step, we report measured generation speedup rather than whole-step refresh
counts.

\begin{table}[h]
\caption{MiniMax H3 controller ablation on a separate 50-step Ref2VA
trajectory. Adaptive variants share the executor and differ only in trigger
cue; Fixed R3 is non-adaptive. Speedup uses the depth-wise executor; PSNR/SSIM
are same-seed diagnostics from the signal-matched closed-loop replay and
indicate trajectory fidelity, not absolute perceptual quality.}
\label{tab:h3-ablation}
\centering
\scriptsize
\setlength{\tabcolsep}{2.0pt}
\renewcommand{\arraystretch}{1.08}
\begin{tabular*}{\linewidth}{@{\extracolsep{\fill}}llccc@{}}
\toprule
Configuration & Controller cue & Speedup $\uparrow$ & PSNR $\uparrow$ & SSIM $\uparrow$ \\
\midrule
Full computation
& -- & 1.000$\times$ & -- & 1.0000 \\

Fixed R3
& fixed interval & 1.464$\times$ & 18.00 & 0.8823 \\

Target-Q
& consecutive target-Q & 1.459$\times$ & 29.81 & 0.9879 \\

Exposure
& directional exposure & 1.459$\times$ & \textbf{32.24} & \textbf{0.9958} \\

\RefAdapt
& target-Q $\times$ exposure & \textbf{1.482$\times$} & 31.24 & 0.9937 \\

Anchor-Q $\times$ Exposure
& anchor-Q $\times$ exposure & 1.376$\times$ & 29.78 & 0.9864 \\

\bottomrule
\end{tabular*}
\end{table}

Across the depth-wise speed ablation and signal-matched fidelity replay in
\cref{tab:h3-ablation}, \RefAdapt provides the best overall operating point:
it attains the highest measured speedup, $1.482\times$, while maintaining
essentially optimal trajectory fidelity (31.24 dB PSNR and 0.9937 SSIM), within
only 1.00 dB and 0.0021 of the numerical maxima and markedly above Fixed R3
(18.00 dB/0.8823). Thus its acceleration does not require leaving the
high-fidelity regime. The two cues are complementary: Target-Q tracks local
target change, exposure tracks whether the target reads Ref, and their product
requires both. The weaker anchor-relative result (1.376$\times$, 29.78
dB/0.9864) further favors local step-to-step change over displacement
accumulated across a variable lag; additional diagnostics are in
\cref{app:observation-details}.

\FloatBarrier

\section{Conclusion}
\label{sec:conclusion}

We propose \RefAdapt, an interaction-conditioned controller that combines
consecutive target-Q change with lagged target-to-reference attention mass to
select a reference reactivation depth, with bounded-age full refresh. On the
fixed 100-case MultiBanana and EditVerseBench sets, it achieves
$2.58\times$ Qwen Image Edit and $1.564\times$ MiniMax H3 generation speedups
while keeping benchmark quality close to full computation. Because ResComp
reduces a different part of the workload than reference-refresh control, the
composed rows' $3.54\times$ and $2.097\times$ speedups provide empirical evidence
of complementarity rather than assuming orthogonality. Thus interaction-aware
reuse removes substantial redundant work even in few-step generation.
Full-length counterfactual analysis identifies the mechanism, while ultra-few-step
closed-loop experiments test realized generation outcomes under a reduced error-recovery
budget. Our evaluation covers two representative joint-attention systems and
benchmark-native image/video tasks; future work will extend \RefAdapt to broader
architectures and sampling regimes, automate controller calibration for new
operating points, and jointly optimize reference reuse with complementary
acceleration techniques.

\section*{Reproducibility Statement}
The method and runtime controller are specified in
\cref{sec:observations,sec:method} and Appendix~\ref{app:method-details}, while
the datasets, distilled checkpoints, hardware, metrics, timing definitions, and
ablations are documented in \cref{sec:experiments} and
Appendix~\ref{app:experiments}.

\section*{AI Use Statement}
Generative AI tools were used only to improve manuscript readability, LaTeX
organization, and language editing. They were not used for research tasks such
as method or experiment design, implementation, data processing, or result
interpretation. All AI-assisted edits were reviewed by the authors, who verified
that they did not alter the technical content and take full responsibility for
the final manuscript.


\appendix

\section{Additional Method Details}
\label{app:method-details}

\subsection{Counterfactual and Exposure Definitions}
\label{app:directional}

We use the same two semantic roles and the same stale-reference counterfactual
as in \cref{sec:obs-exposure,sec:obs-utility}: current target queries and
current target-side K/V are held fixed, while only reference-side K/V are
replaced by values from an earlier refresh anchor. Stale-$K$ and stale-$V$
diagnostics replace one component at a time. Target exposure is the
post-softmax mass from target queries to reference keys defined in
\cref{eq:directional-exposure}; reference-query mass on target keys is reported
only as a structural diagnostic and is not used as a proxy for target
dependence on the reference. The deployed controller uses the one-step-lagged
realized runtime mass in \cref{eq:factorized-risk} rather than the current
full-fresh exposure.

\subsection{Runtime-Signal Selection and Calibration}
\label{app:additional-controller}

The mechanism analysis uses current full-fresh exposure, whereas the deployed
controller uses a lagged mass that was already observed in an earlier forward.
For reference, the probe also measures a same-step diagnostic in which only the
reference keys are replaced by their anchor values,
\begin{align}
M_{t,b,h}^{\TgtTok\rightarrow\RefTok}
&=\operatorname{Exposure}(Q_{\TgtTok}^{t,b,h},K_{\RefTok}^{t,b,h},K_{\TgtTok}^{t,b,h}),\\
M_{t\mid a,b,h}^{\mathrm{cached},\TgtTok\rightarrow\RefTok}
&=\operatorname{Exposure}(Q_{\TgtTok}^{t,b,h},K_{\RefTok}^{a,b,h},K_{\TgtTok}^{t,b,h}).
\end{align}
For the current H3 short-lag probe, these are nearly identical (Pearson $0.9989$,
Spearman $0.9991$, mean absolute difference $0.00298$). This comparison isolates fresh-vs-cached keys at the \emph{same}
step; it should not be confused with the deployed one-step lag
$M^{\TgtTok\rightarrow\RefTok}_{t-1,b,h}$. We therefore treat the deployed lagged mass as a heuristic interaction cue rather
than as an exact surrogate for current full-fresh exposure.

The current measured target-side proxy is consecutive post-RoPE target-Q change.
For comparison, change to the most recent materialization anchor is
\begin{equation}
c^{\mathrm{anchor}}_{t\mid a_b,b,h}=
\frac{\norm{Q_{\TgtTok}^{t,b,h}-Q_{\TgtTok}^{a_b,b,h}}_F}
{\norm{Q_{\TgtTok}^{a_b,b,h}}_F+\eps}.
\end{equation}
On the current H3 trajectory, consecutive change is the stronger candidate in
the existing aligned next-step impact diagnostic. Unlike consecutive change,
anchor-relative change accumulates displacement from the most recent
block-specific materialization anchor, whose age can vary across both steps and
blocks. Under a shared controller threshold, the two signals therefore need not
produce the same refresh schedule or speedup. We use consecutive change in the
main method and leave target K/V, hidden-state, and anchor-change variants to
ablation rather than treating them as equivalent by definition.

The only aggregation required by the main policy is within a block. We use the
same mean over heads as in the main text,
\begin{equation}
\mathcal H(\{g_{t,b,h}\}_{h=1}^{H})
=\frac{1}{H}\sum_{h=1}^{H}g_{t,b,h}.
\label{eq:candidate-head-aggregate}
\end{equation}
The reactivation threshold $\theta$ and maximum cache age $\Delta_{\max}$ are
fixed controller hyperparameters. No cross-block maximum is needed for the
depth-cutoff executor because blocks are evaluated sequentially and the first
threshold crossing ends reuse for the remaining suffix.

\subsection{Algorithm}
\label{app:algorithm}

\begin{algorithm}[H]
\caption{RefAdapt-DiT: Monotone Depth-Adaptive Reference Reuse}
\label{alg:refadapt}
\footnotesize
\begin{algorithmic}[1]
\STATE Materialize step 0; initialize $\mathcal Z_b$ by \cref{eq:block-cache-state}; set $a^{\mathrm{full}}\leftarrow0$
\STATE Store sampled target queries and directional masses for the next step
\FOR{denoising step $t=1,\ldots,N_{\mathrm{step}}-1$}
    \IF{$t-a^{\mathrm{full}}\ge \Delta_{\max}$}
        \STATE Fully recompute the reference stream from block 0; refresh all $\mathcal Z_b$ and set $a^{\mathrm{full}}\leftarrow t$
        \STATE Record target Q and realized directional masses for the next step
        \STATE \textbf{continue}
    \ENDIF
    \STATE $\textsc{reactivated}\leftarrow\mathrm{False}$
    \FOR{block $b=0,\ldots,B-1$}
        \IF{$\textsc{reactivated}=\mathrm{False}$}
            \STATE Form current target Q and compute $D_{t,b}(Q^{\TgtTok,h})$ from consecutive target-Q change
            \STATE Read $M^{\TgtTok\rightarrow\RefTok}_{t-1,b,h}$ from the previous executed attention
            \STATE Compute $g_{t,b,h}=D_{t,b}(Q^{\TgtTok,h})M^{\TgtTok\rightarrow\RefTok}_{t-1,b,h}$ and $\bar g_{t,b}=\mathcal H(\{g_{t,b,h}\}_h)$
            \IF{$\bar g_{t,b}<\theta$}
                \STATE Skip reference rows at block $b$; target attends to $(\bar K_b^{\RefTok},\bar V_b^{\RefTok})$
            \ELSE
                \STATE Target still uses block-$b$ cached reference K/V
                \STATE Restore $\bar H_b^{\RefTok}$ and run reference-query computation with the current target
                \STATE Store the reference output as $\bar H_{b+1}^{\RefTok}$
                \STATE $\textsc{reactivated}\leftarrow\mathrm{True}$
            \ENDIF
        \ELSE
            \STATE Form reference K/V and run the original joint-attention block
            \STATE Update $\bar K_b^{\RefTok},\bar V_b^{\RefTok}$, store $\bar H_{b+1}^{\RefTok}$, and set $a_b\leftarrow t$
        \ENDIF
        \STATE Record the current target Q and realized runtime mass $M^{\TgtTok\rightarrow\RefTok}_{t,b,h}$ for the next step
    \ENDFOR
\ENDFOR
\end{algorithmic}
\end{algorithm}

Except for the bounded-age fallback that performs a full refresh after $\Delta_{\max}$ steps without a block-0 full refresh, the algorithm uses at most one monotone switch per denoising step. Before the boundary,
reference rows are absent and cached K/V are reused. At the boundary, the most
recently materialized reference hidden state is restored and the reference query
is evaluated against the current target; this updates the reference hidden state
for the next block. All deeper blocks then form new reference K/V from that
reactivated stream and refresh their caches. As a result, different blocks
naturally carry different anchors $a_b$. The suffix is not identical to a
full-fresh forward from block 0 because its starting boundary can be stale; this
approximation is included in all reported results.

\subsection{Computational Accounting}
\label{app:compute}

A full joint-attention block forms four score regions:
$Q_{\RefTok}K_{\RefTok}^{\top}$,
$Q_{\RefTok}K_{\TgtTok}^{\top}$,
$Q_{\TgtTok}K_{\RefTok}^{\top}$, and
$Q_{\TgtTok}K_{\TgtTok}^{\top}$. On a cached step, the two reference-query
regions are removed. Ignoring head dimensions, score computation changes from
\begin{equation}
\mathcal O\!\left((N_{\RefTok}+N_{\TgtTok})^2\right)
\quad\text{to}\quad
\mathcal O\!\left(N_{\TgtTok}(N_{\RefTok}+N_{\TgtTok})\right),
\end{equation}
removing approximately
$\mathcal O(N_{\RefTok}(N_{\RefTok}+N_{\TgtTok}))$ score work. Reference-side QKV
projections and other reference-stream operations can also be skipped because the
layer-wise reference K/V are cached. These FLOP reductions are theoretical; practical
speedups are reported only from measured generation latency under matched software and
hardware settings.

\section{Additional Experimental Details}
\label{app:experiments}

\subsection{Evaluation Protocol}

For Qwen Image Edit, we evaluate a fixed 100-case evaluation set from
MultiBanana-Benchmark; for MiniMax H3, we evaluate a fixed 100-case evaluation
set from EditVerseBench. All runs use NVIDIA H20 GPUs with 96 GB memory per GPU.
Qwen uses the distilled 8-step Image Edit checkpoint at $1024\times1024$, seed
1234, and one GPU. H3 uses the LightX2V release of the distilled 4-step MiniMax
H3 checkpoint at $1344\times768$ (768p), seed 42, and the same GPU type. Within
each setting, model, sampler, precision, batch size, and hardware are identical
across methods. The separate 50-step replay in \cref{sec:exp-ablation} is a
controller ablation, not the 4-step main benchmark. Generation latency is
measured from real runs; theoretical FLOPs and kernel timings, if reported, are
kept separate from measured latency.

\subsection{Metrics and Baseline Configuration}

For Qwen, the baseline set is TeaCache, SeaCache, TaylorSeer, and ToMe. For
MiniMax H3, we use the same set and additionally include AsymRnR. The composed
row in each main table combines \RefAdapt with the same orthogonal
resolution-compression optimization and is reported separately from the
standalone method.

For MultiBanana, we run the benchmark-provided Qwen3-VL-8B-Instruct judge and
retain its five 1--10 scores---instruction alignment, reference consistency,
background--subject match, physical realism, and visual quality---together with
the benchmark's official weighted score
$3\,\mathrm{IA}+3\,\mathrm{RC}+\mathrm{BSM}+\mathrm{PR}+\mathrm{VQ}$.
For EditVerseBench, we follow the released evaluation fields: VLM editing
quality, Pick Score video quality, frame-level text alignment, video-level text
alignment, CLIP temporal consistency, and DINO temporal consistency. We do not
collapse the EditVerse metrics into a custom aggregate.

The main tables use generation speedup relative to full computation as
the efficiency metric. TaylorSeer uses interval 2 in the Qwen main table. The
standalone \RefAdapt rows use the adaptive reference-reuse controller studied
in the method section.

\subsection{Qwen Performance by Number of References}

Table~\ref{tab:qwen-refcount-speed} reports the generation speedup as the
number of reference images increases. This breakdown is supplementary to the
100-case aggregate in Table~\ref{tab:image-main} and is not used as a separate
quality claim.

\begin{table}[h]
\caption{Qwen generation speedup by number of reference images. TaylorSeer
uses interval 2. Speedup is relative to full computation at the same reference
count.}
\label{tab:qwen-refcount-speed}
\centering
\small
\setlength{\tabcolsep}{3.5pt}
\begin{tabular}{@{}lrrrrrrr@{}}
\toprule
Method & 3 refs & 4 refs & 5 refs & 6 refs & 7 refs & 8 refs & Overall \\
\midrule
ToMe & 1.09$\times$ & 1.08$\times$ & 1.07$\times$ & 1.07$\times$ & 1.06$\times$ & 1.07$\times$ & 1.06$\times$ \\
TeaCache & 1.13$\times$ & 1.14$\times$ & 1.13$\times$ & 1.14$\times$ & 1.13$\times$ & 1.13$\times$ & 1.13$\times$ \\
SeaCache & 1.59$\times$ & 1.60$\times$ & 1.60$\times$ & 1.60$\times$ & 1.57$\times$ & 1.62$\times$ & 1.60$\times$ \\
TaylorSeer ($I{=}2$) & 1.59$\times$ & 1.60$\times$ & 1.60$\times$ & 1.60$\times$ & 1.60$\times$ & 1.61$\times$ & 1.61$\times$ \\
\RefAdapt & \textbf{2.09$\times$} & \textbf{2.26$\times$} & \textbf{2.38$\times$} & \textbf{2.49$\times$} & \textbf{2.57$\times$} & \textbf{2.62$\times$} & \textbf{2.58$\times$} \\
\bottomrule
\end{tabular}
\end{table}

\section{Additional Quantitative Analysis}
\label{app:observation-details}

\subsection{Mechanism Statistics for H3 and Qwen}
\label{app:h3-qwen-mechanism}

Table~\ref{tab:h3-qwen-mechanism} collects the compact mechanism statistics
used in the main-text H3/Qwen comparison. The two settings use different raw
token partitions within the shared reference/target roles, so absolute values
are reported within each setting rather than pooled across them.

\begin{table}[h]
\caption{Mechanism statistics for the current canonical H3 and Qwen cases.
Spearman correlations are within-case diagnostics and are reported separately
for each setting.}
\label{tab:h3-qwen-mechanism}
\centering
\small
\begin{tabular}{lcc}
\toprule
Diagnostic & H3 Ref2VA & Qwen multi-image \\
\midrule
Raw endpoint vs. $I^{KV}$ & 0.564 & 0.708 \\
Endpoint $\times$ exposure vs. $I^{KV}$ & 0.826 & 0.905 \\
Target-Q change vs. $I^{KV}_{t+1}$ & 0.607 & 0.710 \\
Target-Q $\times$ runtime exposure vs. $I^{KV}_{t+1}$ & 0.869 & 0.846 \\
Sensitivity blocks improved by exposure & 9/9 & 7/7 \\
\bottomrule
\end{tabular}
\end{table}

Here $I^{KV}$ denotes the local target-output perturbation produced by replacing
current reference-side K/V with stale values from the relevant refresh anchor
while holding the target side fixed. ``Within-case'' means that each Spearman
coefficient is computed across aligned internal measurements (e.g., blocks,
heads, and anchors) from one canonical trajectory, rather than across
independent benchmark inputs. These numbers are therefore mechanism and signal
selection diagnostics, not population-level estimates of benchmark behavior.

For Qwen, reversing the order of the same two reference images gives a raw
endpoint-to-impact Spearman of $0.685$ and an exposure-weighted value of $0.911$.
The control also shows that the two reference images are not interchangeable:
their exposure and individual stale impact differ even when the aggregate
reference-side behavior remains similar.

\subsection{Reference--Target Dynamics}

In the current full video trajectory, the median reference-drift/target-drift
ratio over matched K/V measurements in blocks 6--25 is $0.352$. The gap narrows
around block 31 and reverses in deeper blocks. Target-state change and
reference-K/V change also share some timing structure: the median per-block
Spearman correlation is $0.675$ across 50 blocks, and 22/50 blocks reach at
least $0.8$. These statistics describe one complete H3 video case and are
reported as within-case diagnostics.

\begin{figure}[!htbp]
    \centering
    \includegraphics[width=\linewidth]{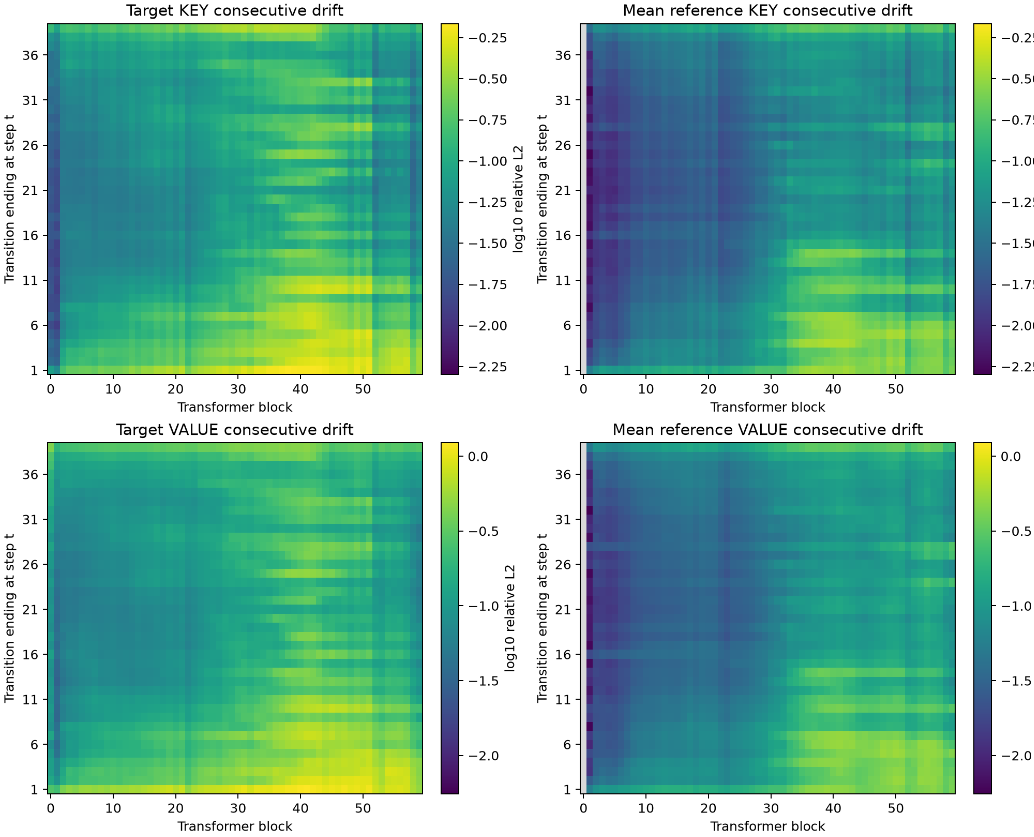}
    \caption{Retained left-half panels from the full-block Qwen reference/target
    dynamics diagnostics. These four panels report target and mean-reference
    consecutive drift for K and V across blocks and denoising steps; the
    dropped right-half panels contained the corresponding log-ratio and
    per-block timing-coupling summaries.}
    \label{fig:app-qwen-full-depth-left}
\end{figure}

\subsection{Directional Attention Diagnostics}

For target-video queries in the current video case, Ref receives $9.51\%$ of
attention mass on average and the remaining $90.49\%$ is assigned to Tgt.
Reference-video queries assign $26.14\%$ mass to target-video keys, versus
$9.51\%$ in the opposite direction; the directional difference is positive in
$97.22\%$ of step--block cells. These values are structural diagnostics and
are not used to infer target dependence from reference-query attention.

\begin{figure}[!htbp]
    \centering
    \includegraphics[width=\linewidth]{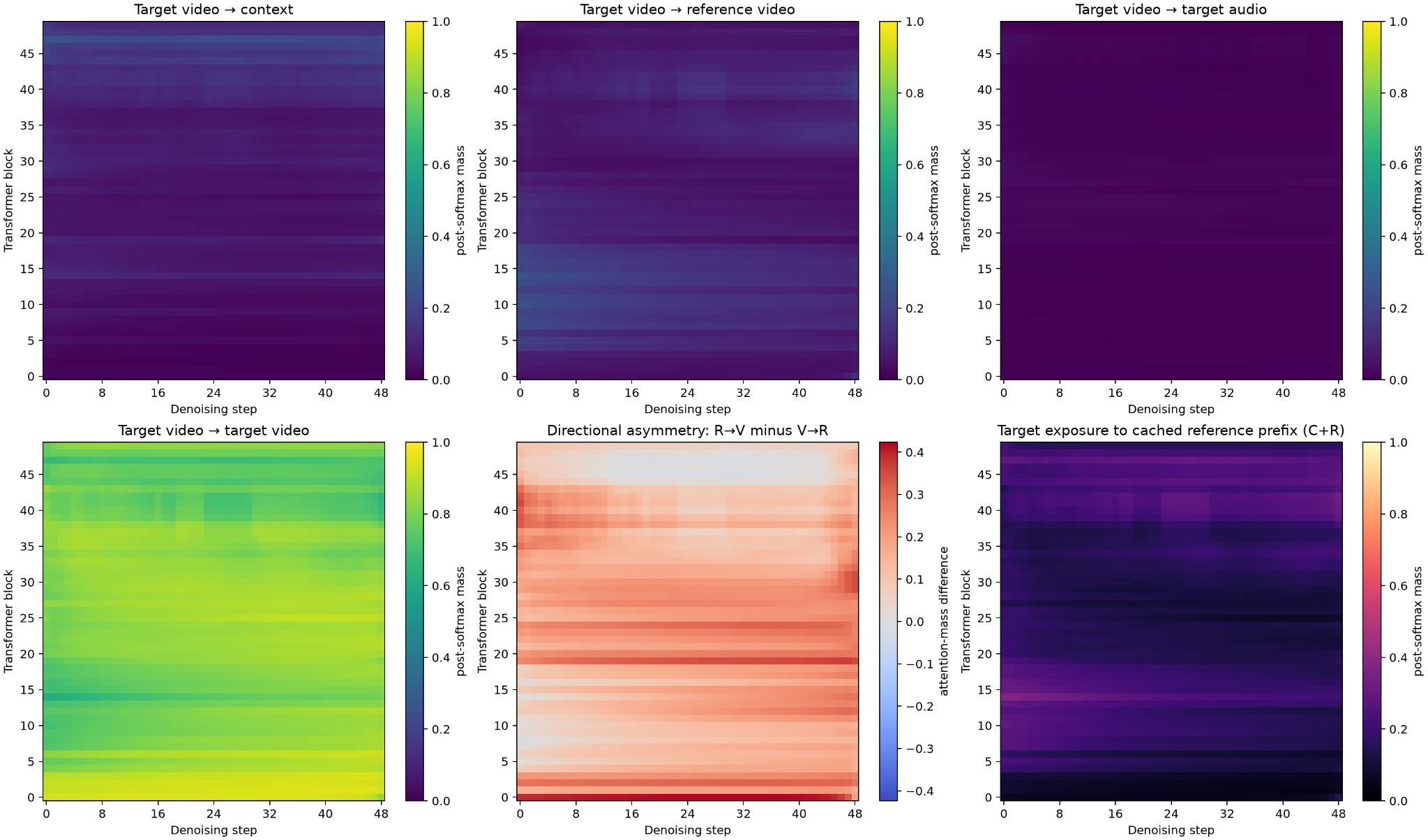}
    \caption{Complete H3 raw-partition attention diagnostics. The main text
    keeps target exposure to Ref and aggregates the remaining mass as Tgt; the
    remaining panels report implementation-level allocation and directional
    asymmetry. Every arrow in the panel titles runs from query role to key
    role, not along causal flow.}
    \label{fig:app-h3-role-atlas}
\end{figure}

\subsection{Runtime-Signal Diagnostics}

Table~\ref{tab:signal-analysis} complements the cross-model summary in
Table~\ref{tab:h3-qwen-mechanism} with H3-only runtime-signal checks that are
not already reported there. In particular, it contrasts the deployed
consecutive-change design with an anchor-relative alternative.

\begin{table}[!htbp]
\caption{Additional H3 runtime-signal diagnostics from one complete Ref2VA
trajectory. Correlations are within-case screening statistics, not confidence
estimates over independent inputs.}
\label{tab:signal-analysis}
\centering
\small
\begin{tabular}{lcc}
\toprule
Signal / target & Pearson $\uparrow$ & Spearman $\uparrow$\\
\midrule
Runtime stale-reference exposure vs. $I^{KV}_{t+1}$ & 0.479 & 0.789\\
Target-Q anchor change vs. $I^{KV}_{t+1}$ & 0.390 & 0.418\\
Anchor change $\times$ exposure vs. $I^{KV}_{t+1}$ & 0.589 & 0.545\\
\bottomrule
\end{tabular}
\end{table}

The anchor-relative signal measures displacement from the most recent
block-specific materialization anchor, which may be several denoising steps old,
whereas the deployed Target-Q signal measures only the consecutive
step-to-step change. The former therefore accumulates motion over a variable
lag and has a different numerical scale and trigger history under the same
controller threshold. On the current ablation, this makes the anchor-relative
variant more conservative about reuse and yields lower measured speedup; this
comparison should be read as a fixed-controller ablation rather than as a
threshold-matched universal ranking of the two signal families.

\begin{figure}[!htbp]
    \centering
    \begin{minipage}[c]{0.48\linewidth}
        \centering
        \includegraphics[width=\linewidth]{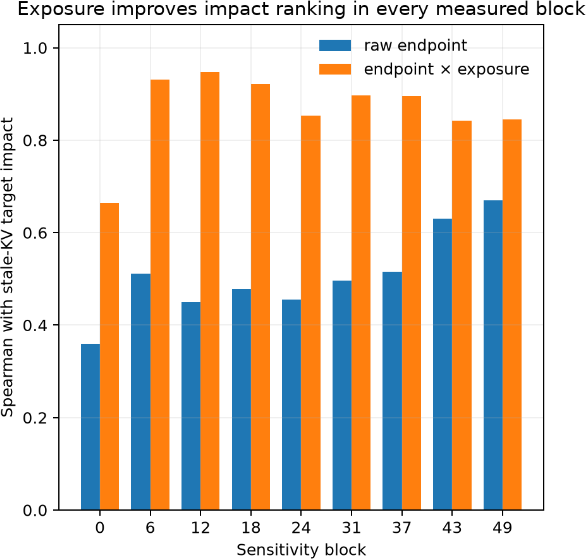}\\[-1pt]
        {\small (a) MiniMax H3}
    \end{minipage}\hfill
    \begin{minipage}[c]{0.48\linewidth}
        \centering
        \includegraphics[width=\linewidth]{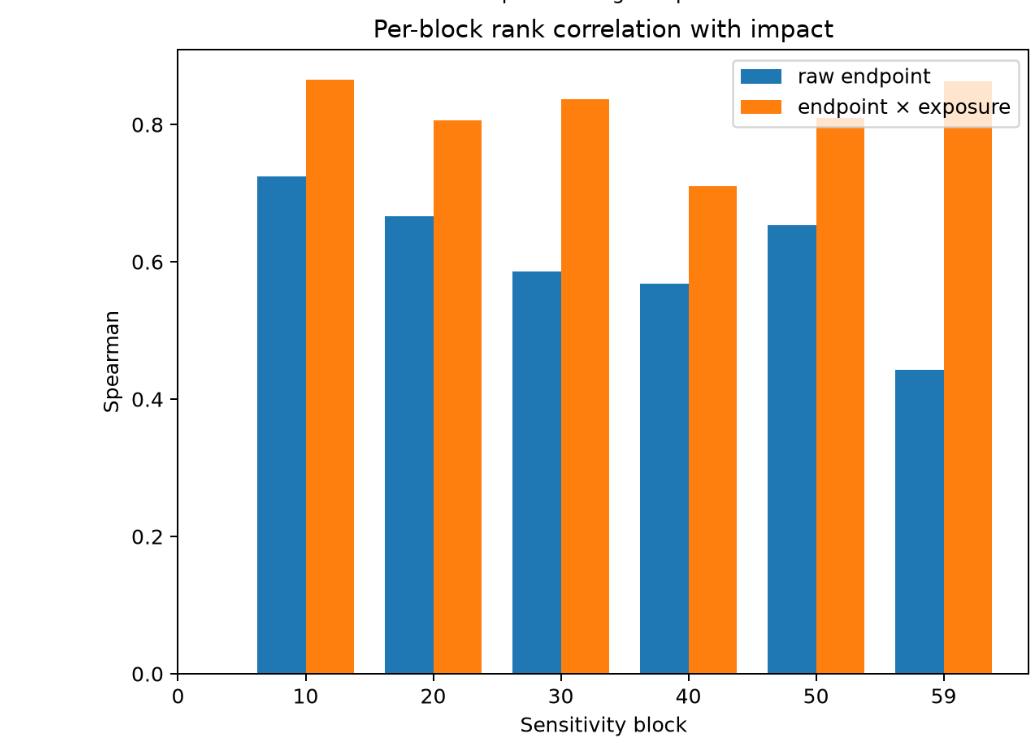}\\[-1pt]
        {\small (b) Qwen Image Edit}
    \end{minipage}
    \caption{Block-wise support for exposure conditioning. In every measured
    sensitivity block, endpoint drift weighted by target exposure has higher
    Spearman association with local stale-KV impact than raw endpoint drift
    (9/9 H3 blocks and 7/7 Qwen blocks).}
    \label{fig:app-impact-ranking}
\end{figure}
\FloatBarrier

\subsection{Probe Non-Intrusiveness and Multi-Reference Controls}

For the canonical Qwen probe, enabling the observation sidecar leaves the final
PNG bit-identical to the probe-off run (matching SHA-256). Probe-on generation
takes 247.1 s versus 232.9 s probe-off, corresponding to roughly 6.1\% overhead
from the optional research logging sidecar. This sidecar is used to collect
additional diagnostic traces and is distinct from the runtime controller; it is
not part of the production configuration whose speedups are reported in the
main text.

\section{Additional Qualitative Results}
\label{app:qualitative}

\subsection{Prompts Used for the Figure~1 Cross-Prompt Profiles}
\label{app:fig1-prompts}

The magnitude profiles in \cref{fig:obs-dynamics} use the following three
prompts per model. Within each model, the reference inputs, seed, resolution,
and denoising configuration are held fixed; only the prompt changes. The
heatmap averages \cref{eq:drift-log-ratio} cellwise across these prompts. The
profile first averages each prompt over denoising transitions and then averages
the prompt-level values as specified in
\cref{eq:cross-prompt-reference-profile}.

\begin{table}[H]
\centering
\small
\setlength{\tabcolsep}{4pt}
\renewcommand{\arraystretch}{1.08}
\begin{tabular}{@{}p{0.13\linewidth}p{0.20\linewidth}p{0.61\linewidth}@{}}
\toprule
Model & Case & Prompt \\
\midrule
MiniMax H3 & \texttt{canonical} & Reimagine the animal from the reference video as a giant panda ambling through the snowy forest, same motion and camera path, breath misting, soft wind and distant birdsong \\
MiniMax H3 & \texttt{red\_fox\_autumn} & Reimagine the animal from the reference video as a red fox trotting through an autumn maple forest, preserve the original motion and camera path, rustling leaves, warm sunset light, and natural forest ambience \\
MiniMax H3 & \texttt{robot\_dog\_neon} & Reimagine the animal from the reference video as a small silver robot dog exploring a rain-soaked neon city street, preserve the original motion and camera path, reflected lights, light rain, and distant traffic ambience \\
\midrule
Qwen & \texttt{canonical} & the woman and man are hugging together \\
Qwen & \texttt{outdoor\_cafe} & Place the woman and the man seated across from each other at a small outdoor cafe table, preserving both identities and realistic facial details \\
Qwen & \texttt{umbrella\_neon} & Place the woman and the man standing back-to-back under one transparent umbrella on a rainy neon-lit street, preserving both identities and realistic facial details \\
\bottomrule
\end{tabular}
\caption{Exact prompts used to construct the three-prompt reference-drift
profiles in \cref{fig:obs-dynamics}.}
\label{tab:fig1-prompts}
\end{table}


\begin{thebibliography}{32}
\providecommand{\natexlab}[1]{#1}
\providecommand{\url}[1]{\texttt{#1}}
\expandafter\ifx\csname urlstyle\endcsname\relax
  \providecommand{\doi}[1]{doi: #1}\else
  \providecommand{\doi}{doi: \begingroup \urlstyle{rm}\Url}\fi

\bibitem[Bian et~al.(2026)Bian, Chen, Li, Zhi, Sang, Luo, and
  Xu]{videoasprompt}
Yuxuan Bian, Xin Chen, Zenan Li, Tiancheng Zhi, Shen Sang, Linjie Luo, and
  Qiang Xu.
\newblock {Video-As-Prompt}: Unified semantic control for video generation.
\newblock In \emph{International Conference on Learning Representations}, 2026.

\bibitem[Bolya \& Hoffman(2023)Bolya and Hoffman]{tome}
Daniel Bolya and Judy Hoffman.
\newblock Token merging for fast stable diffusion.
\newblock In \emph{Proceedings of the IEEE/CVF Conference on Computer Vision
  and Pattern Recognition Workshops}, pp.\  4599--4603, 2023.

\bibitem[Bu et~al.(2026)Bu, Ling, Zhou, Wang, Zang, Lin, and Wang]{dicache}
Jiazi Bu, Pengyang Ling, Yujie Zhou, Yibin Wang, Yuhang Zang, Dahua Lin, and
  Jiaqi Wang.
\newblock {DiCache}: Let diffusion model determine its own cache.
\newblock In \emph{International Conference on Learning Representations}, 2026.

\bibitem[Chen et~al.(2026{\natexlab{a}})Chen, Zheng, Lin, and Zhang]{svdcache}
Guantao Chen, Shikang Zheng, Yuqi Lin, and Linfeng Zhang.
\newblock Forecast the principal, stabilize the residual: Subspace-aware
  feature caching for diffusion transformers.
\newblock In \emph{Proceedings of the IEEE/CVF Conference on Computer Vision
  and Pattern Recognition}, pp.\  23632--23641, 2026{\natexlab{a}}.

\bibitem[Chen et~al.(2026{\natexlab{b}})Chen, Zeng, Zhao, Shen, Ye, Xiang,
  Wang, Cheng, Yu, and Chen]{regione}
Pengtao Chen, Xianfang Zeng, Maosen Zhao, Mingzhu Shen, Peng Ye, Bangyin Xiang,
  Zhibo Wang, Wei Cheng, Gang Yu, and Tao Chen.
\newblock {RegionE}: Adaptive region-aware generation for efficient image
  editing.
\newblock In \emph{International Conference on Learning Representations},
  2026{\natexlab{b}}.

\bibitem[Chen et~al.(2025)Chen, Zhang, Zhang, Zhou, Kim, Liu, Li, Zhang, Zhao,
  Wang, Ding, Lin, and Zhao]{unireal}
Xi~Chen, Zhifei Zhang, He~Zhang, Yuqian Zhou, Soo~Ye Kim, Qing Liu, Yijun Li,
  Jianming Zhang, Nanxuan Zhao, Yilin Wang, Hui Ding, Zhe Lin, and Hengshuang
  Zhao.
\newblock {UniReal}: Universal image generation and editing via learning
  real-world dynamics.
\newblock In \emph{Proceedings of the IEEE/CVF Conference on Computer Vision
  and Pattern Recognition}, pp.\  12501--12511, 2025.

\bibitem[Chu et~al.(2025)Chu, Wu, Feng, and Zhang]{omnicache}
Huanpeng Chu, Wei Wu, Guanyu Feng, and Yutao Zhang.
\newblock {OmniCache}: A trajectory-oriented global perspective on
  training-free cache reuse for diffusion transformer models.
\newblock In \emph{Proceedings of the IEEE/CVF International Conference on
  Computer Vision}, pp.\  16302--16312, 2025.

\bibitem[Chung et~al.(2026)Chung, Hyun, Lee, Han, Cha, Wee, Hong, and
  Heo]{seacache}
Jiwoo Chung, Sangeek Hyun, MinKyu Lee, Byeongju Han, Geonho Cha, Dongyoon Wee,
  Youngjun Hong, and Jae-Pil Heo.
\newblock {SeaCache}: Spectral-evolution-aware cache for accelerating diffusion
  models.
\newblock In \emph{Proceedings of the IEEE/CVF Conference on Computer Vision
  and Pattern Recognition}, pp.\  14283--14294, 2026.

\bibitem[Cui et~al.(2026)Cui, Tang, Xu, Yao, Zeng, and Jia]{bwcache}
Hanshuai Cui, Zhiqing Tang, Zhifei Xu, Zhi Yao, Wenyi Zeng, and Weijia Jia.
\newblock {BWCache}: Accelerating video diffusion transformers through
  block-wise caching.
\newblock In \emph{International Conference on Learning Representations}, 2026.

\bibitem[Gu et~al.(2026)Gu, He, Su, He, Wang, and Liu]{scalingcache}
Lihui Gu, Jingbin He, Lianghao Su, Kang He, Wenxiao Wang, and Yuliang Liu.
\newblock {ScalingCache}: Extreme acceleration of {DiTs} through difference
  scaling and dynamic interval caching.
\newblock In \emph{International Conference on Learning Representations}, 2026.

\bibitem[Ju et~al.(2026)Ju, Wang, Zhou, Zhang, Liu, Zhao, Zhang, Li, Cai, Liu,
  Pakhomov, Lin, Kim, and Xu]{editverse}
Xuan Ju, Tianyu Wang, Yuqian Zhou, He~Zhang, Qing Liu, Cherry Zhao, Zhifei
  Zhang, Yijun Li, Yuanhao Cai, Shaoteng Liu, Daniil Pakhomov, Zhe Lin, Soo~Ye
  Kim, and Qiang Xu.
\newblock {EditVerse}: Unifying image and video editing and generation with
  in-context learning.
\newblock In \emph{International Conference on Learning Representations}, 2026.

\bibitem[Kahatapitiya et~al.(2025)Kahatapitiya, Liu, He, Liu, Jia, Zhang, Ryoo,
  and Xie]{adacache}
Kumara Kahatapitiya, Haozhe Liu, Sen He, Ding Liu, Menglin Jia, Chenyang Zhang,
  Michael~S. Ryoo, and Tian Xie.
\newblock Adaptive caching for faster video generation with diffusion
  transformers.
\newblock In \emph{Proceedings of the IEEE/CVF International Conference on
  Computer Vision}, pp.\  15240--15252, 2025.

\bibitem[Liu et~al.(2025{\natexlab{a}})Liu, Zhang, Wang, Wei, Qiu, Zhao, Zhang,
  Ye, and Wan]{teacache}
Feng Liu, Shiwei Zhang, Xiaofeng Wang, Yujie Wei, Haonan Qiu, Yuzhong Zhao,
  Yingya Zhang, Qixiang Ye, and Fang Wan.
\newblock Timestep embedding tells: It's time to cache for video diffusion
  model.
\newblock In \emph{Proceedings of the IEEE/CVF Conference on Computer Vision
  and Pattern Recognition}, pp.\  7353--7363, 2025{\natexlab{a}}.

\bibitem[Liu et~al.(2026)Liu, Cheng, Miao, Liu, Chen, Lin, Yao, Chen, Leng,
  Guo, and Feng]{astraea}
Haosong Liu, Yuge Cheng, Wenxuan Miao, Zihan Liu, Aiyue Chen, Jing Lin, Yiwu
  Yao, Chen Chen, Jingwen Leng, Minyi Guo, and Yu~Feng.
\newblock {Astraea}: A token-wise acceleration framework for video diffusion
  transformers.
\newblock In \emph{International Conference on Learning Representations}, 2026.

\bibitem[Liu et~al.(2025{\natexlab{b}})Liu, Zou, Lyu, Chen, and
  Zhang]{taylorseer}
Jiacheng Liu, Chang Zou, Yuanhuiyi Lyu, Junjie Chen, and Linfeng Zhang.
\newblock From reusing to forecasting: Accelerating diffusion models with
  {TaylorSeers}.
\newblock In \emph{Proceedings of the IEEE/CVF International Conference on
  Computer Vision}, pp.\  15853--15863, 2025{\natexlab{b}}.

\bibitem[Lyu et~al.(2025)Lyu, Si, Song, Yang, Qiao, Liu, and Wong]{fastercache}
Zhengyao Lyu, Chenyang Si, Junhao Song, Zhenyu Yang, Yu~Qiao, Ziwei Liu, and
  Kwan-Yee~K. Wong.
\newblock {FasterCache}: Training-free video diffusion model acceleration with
  high quality.
\newblock In \emph{International Conference on Learning Representations}, 2025.

\bibitem[Ma et~al.(2024)Ma, Fang, and Wang]{deepcache}
Xinyin Ma, Gongfan Fang, and Xinchao Wang.
\newblock {DeepCache}: Accelerating diffusion models for free.
\newblock In \emph{Proceedings of the IEEE/CVF Conference on Computer Vision
  and Pattern Recognition}, pp.\  15762--15772, 2024.

\bibitem[Oshima et~al.(2026)Oshima, Miyake, Matsutani, Iwasawa, Suzuki, Matsuo,
  and Furuta]{multibanana}
Yuta Oshima, Daiki Miyake, Kohsei Matsutani, Yusuke Iwasawa, Masahiro Suzuki,
  Yutaka Matsuo, and Hiroki Furuta.
\newblock {MultiBanana}: A challenging benchmark for multi-reference
  text-to-image generation.
\newblock In \emph{Proceedings of the IEEE/CVF Conference on Computer Vision
  and Pattern Recognition}, pp.\  448--460, 2026.

\bibitem[Peebles \& Xie(2023)Peebles and Xie]{dit}
William Peebles and Saining Xie.
\newblock Scalable diffusion models with transformers.
\newblock In \emph{Proceedings of the IEEE/CVF International Conference on
  Computer Vision}, pp.\  4195--4205, 2023.

\bibitem[Son et~al.(2026)Son, Jeon, Choi, and Ham]{rfc}
Byunggwan Son, Jeimin Jeon, Jeongwoo Choi, and Bumsub Ham.
\newblock Relational feature caching for accelerating diffusion transformers.
\newblock In \emph{International Conference on Learning Representations}, 2026.

\bibitem[Sun et~al.(2025)Sun, Tu, Liao, Jin, and Tao]{asymrnr}
Wenhao Sun, Rong-Cheng Tu, Jingyi Liao, Zhao Jin, and Dacheng Tao.
\newblock {AsymRnR}: Video diffusion transformers acceleration with asymmetric
  reduction and restoration.
\newblock In \emph{Proceedings of the 42nd International Conference on Machine
  Learning}, pp.\  57694--57711, 2025.

\bibitem[Tan et~al.(2026)Tan, Xue, Yang, Liu, and Wang]{ominicontrol2}
Zhenxiong Tan, Qiaochu Xue, Xingyi Yang, Songhua Liu, and Xinchao Wang.
\newblock {OminiControl2}: Efficient conditioning for diffusion transformers.
\newblock In \emph{Proceedings of the IEEE/CVF Conference on Computer Vision
  and Pattern Recognition}, pp.\  4256--4265, 2026.

\bibitem[Wimbauer et~al.(2024)Wimbauer, Wu, Schoenfeld, Dai, Hou, He,
  Sanakoyeu, Zhang, Tsai, Kohler, Rupprecht, Cremers, Vajda, and Wang]{cacheme}
Felix Wimbauer, Bichen Wu, Edgar Schoenfeld, Xiaoliang Dai, Ji~Hou, Zijian He,
  Artsiom Sanakoyeu, Peizhao Zhang, Sam Tsai, Jonas Kohler, Christian
  Rupprecht, Daniel Cremers, Peter Vajda, and Jialiang Wang.
\newblock Cache me if you can: Accelerating diffusion models through block
  caching.
\newblock In \emph{Proceedings of the IEEE/CVF Conference on Computer Vision
  and Pattern Recognition}, pp.\  6211--6220, 2024.

\bibitem[Wu et~al.(2025)Wu, Li, Zhou, Lin, Gao, Yan, Yin, Bai, Xu, Chen, Chen,
  Tang, Zhang, Wang, Yang, Yu, Cheng, Liu, Li, Zhang, Meng, Wei, Ni, Chen, Cao,
  Peng, Qu, Wu, Wang, Yu, Wen, Feng, Xu, Wang, Zhang, Zhu, Wu, Cai, and
  Liu]{qwenimage}
Chenfei Wu, Jiahao Li, Jingren Zhou, Junyang Lin, Kaiyuan Gao, Kun Yan,
  Sheng-ming Yin, Shuai Bai, Xiao Xu, Yilei Chen, Yuxiang Chen, Zecheng Tang,
  Zekai Zhang, Zhengyi Wang, An~Yang, Bowen Yu, Chen Cheng, Dayiheng Liu,
  Deqing Li, Hang Zhang, Hao Meng, Hu~Wei, Jingyuan Ni, Kai Chen, Kuan Cao,
  Liang Peng, Lin Qu, Minggang Wu, Peng Wang, Shuting Yu, Tingkun Wen, Wensen
  Feng, Xiaoxiao Xu, Yi~Wang, Yichang Zhang, Yongqiang Zhu, Yujia Wu, Yuxuan
  Cai, and Zenan Liu.
\newblock {Qwen-Image} technical report, 2025.

\bibitem[Wu et~al.(2026)Wu, Hou, Yang, Tian, Wan, Zhang, and Tong]{vmoba}
Jianzong Wu, Liang Hou, Haotian Yang, Ye~Tian, Pengfei Wan, Di~Zhang, and
  Yunhai Tong.
\newblock {VMoBA}: Mixture-of-block attention for video diffusion models.
\newblock In \emph{International Conference on Learning Representations}, 2026.

\bibitem[Xia et~al.(2025)Xia, Zhang, Li, Wang, Wang, Wu, Yu, and
  Jia]{dreamomni}
Bin Xia, Yuechen Zhang, Jingyao Li, Chengyao Wang, Yitong Wang, Xinglong Wu,
  Bei Yu, and Jiaya Jia.
\newblock {DreamOmni}: Unified image generation and editing.
\newblock In \emph{Proceedings of the IEEE/CVF Conference on Computer Vision
  and Pattern Recognition}, pp.\  28533--28543, 2025.

\bibitem[Xiao et~al.(2025)Xiao, Wang, Zhou, Yuan, Xing, Yan, Li, Wang, Huang,
  and Liu]{omnigen}
Shitao Xiao, Yueze Wang, Junjie Zhou, Huaying Yuan, Xingrun Xing, Ruiran Yan,
  Chaofan Li, Shuting Wang, Tiejun Huang, and Zheng Liu.
\newblock {OmniGen}: Unified image generation.
\newblock In \emph{Proceedings of the IEEE/CVF Conference on Computer Vision
  and Pattern Recognition}, pp.\  13294--13304, 2025.

\bibitem[Ye et~al.(2026)Ye, He, Liu, Wang, Wang, Wan, Zhang, Gai, Chen, and
  Luo]{unic}
Zixuan Ye, Xuanhua He, Quande Liu, Qiulin Wang, Xintao Wang, Pengfei Wan,
  Di~Zhang, Kun Gai, Qifeng Chen, and Wenhan Luo.
\newblock Unified in-context video editing.
\newblock In \emph{International Conference on Learning Representations}, 2026.

\bibitem[Zhang et~al.(2025{\natexlab{a}})Zhang, Xing, Xia, Liu, Peng, Tao, Wan,
  Lo, and Jia]{jenga}
Yuechen Zhang, Jinbo Xing, Bin Xia, Shaoteng Liu, Bohao Peng, Xin Tao, Pengfei
  Wan, Eric Lo, and Jiaya Jia.
\newblock Training-free efficient video generation via dynamic token carving.
\newblock In \emph{Advances in Neural Information Processing Systems},
  volume~38, 2025{\natexlab{a}}.

\bibitem[Zhang et~al.(2025{\natexlab{b}})Zhang, Yuan, Song, Wang, and
  Liu]{easycontrol}
Yuxuan Zhang, Yirui Yuan, Yiren Song, Haofan Wang, and Jiaming Liu.
\newblock {EasyControl}: Adding efficient and flexible control for diffusion
  transformer.
\newblock In \emph{Proceedings of the IEEE/CVF International Conference on
  Computer Vision}, pp.\  19513--19524, 2025{\natexlab{b}}.

\bibitem[Zhao et~al.(2025)Zhao, Jin, Wang, and You]{pab}
Xuanlei Zhao, Xiaolong Jin, Kai Wang, and Yang You.
\newblock Real-time video generation with pyramid attention broadcast.
\newblock In \emph{International Conference on Learning Representations}, 2025.

\bibitem[Zheng et~al.(2026)Zheng, Chen, Zhou, Lin, He, Zou, Cai, Liu, and
  Zhang]{hyca}
Shikang Zheng, Guantao Chen, Qinming Zhou, Yuqi Lin, Lixuan He, Chang Zou,
  Peiliang Cai, Jiacheng Liu, and Linfeng Zhang.
\newblock Let features decide their own solvers: Hybrid feature caching for
  diffusion transformers.
\newblock In \emph{International Conference on Learning Representations}, 2026.

\end{thebibliography}
\end{document}